%% file: icml2026.tex
\documentclass{article}

\usepackage{microtype}
\usepackage{graphicx}
\usepackage{subcaption}
\usepackage{booktabs} 
\usepackage{enumitem}

\usepackage{hyperref}

\usepackage[preprint]{icml2026}

\input{math_commands.tex}


\usepackage{wrapfig} 
\usepackage{hyperref}
\usepackage{url}
\definecolor{iccvblue}{rgb}{0.21,0.49,0.74}
\usepackage{tabularx}    
\usepackage{makecell}    
\usepackage{tcolorbox}
\usepackage{multirow}
\definecolor{verylightgray}{HTML}{F8F8F8}
\usepackage{xcolor}
\usepackage{colortbl}
\usepackage{booktabs}
\usepackage{adjustbox}
\definecolor{mypink}{HTML}{FBF3DA}
\newcommand{\shortname}{MC-LLaVA}

\usepackage{amsmath}
\usepackage{amssymb}
\usepackage{mathtools}
\usepackage{amsthm}

\usepackage[capitalize,noabbrev]{cleveref}

\theoremstyle{plain}

\theoremstyle{definition}

\theoremstyle{remark}

\usepackage[textsize=tiny]{todonotes}

\icmltitlerunning{MC-LLaVA: Multi Concept Personalized Vision Language Model}

\begin{document}

\twocolumn[
  \icmltitle{MC-LLaVA: Multi Concept Personalized Vision Language Model}


  \icmlsetsymbol{equal}{*}
  \icmlsetsymbol{corr}{‡}

  \begin{icmlauthorlist}
    \icmlauthor{Ruichuan An}{equal,pku}
    \icmlauthor{Sihan Yang}{equal,pku}
    \icmlauthor{Renrui Zhang}{cuhk}
    \icmlauthor{Ming Lu}{intel}
    \icmlauthor{Tianyi Jiang}{pku}
    \icmlauthor{Kai Zeng}{pku}
    \icmlauthor{Yulin Luo}{pku}
    \\
    \icmlauthor{Jiajun Cao}{pku}
    \icmlauthor{Hao Liang}{pku}
    \icmlauthor{Ying Chen}{bd}
    \icmlauthor{Qi She}{bd}
    \icmlauthor{Shanghang Zhang}{corr,pku}
    \icmlauthor{Wentao Zhang}{corr,pku}
  \end{icmlauthorlist}

  \icmlaffiliation{pku}{Peking University}
  \icmlaffiliation{cuhk}{CUHK}
  \icmlaffiliation{bd}{ByteDance}
  \icmlaffiliation{intel}{Intel Labs, China}

  \icmlcorrespondingauthor{Shanghang Zhang and Wentao Zhang}{wentao.zhang@pku.edu.cn}

  \icmlkeywords{Vision Language Models, Personalization}

  \vskip 0.3in
]



\printAffiliationsAndNotice{\icmlEqualContribution\icmlCorr}
\input{sec/0_abstract}
\input{sec/1_intro}
\input{sec/2_related_work}

\input{sec/3_method}
\input{sec/4_dataset}

\input{sec/5_exp}
\input{sec/6_conclusion}
\newpage
\section*{Impact Statement}
In this work, we propose MC-LLaVA, a pioneering framework and benchmark aimed at propelling the field of multi-concept personalization within vision-language models. Through a structured methodology for joint concept learning and the introduction of a premium instruction dataset, we facilitate the creation of robust, user-centric AI agents adept at parsing complex, multi-subject environments. Our contribution to multimodal personalization carries positive societal implications, promising the development of more intuitive and accessible tools for sectors like personalized education and creative content curation. Notably, by leveraging a character-centric dataset sourced from films, we advance personalization research while effectively circumventing the privacy risks often linked to personal photographic data. We anticipate no specific immediate negative societal outcomes, barring well-known challenges inherent to machine learning, such as potential biases in foundation models.

\bibliography{icml2026}
\bibliographystyle{icml2026}

\appendix
\onecolumn
\input{sec/X_suppl}

\end{document}

%% file: math_commands.tex
\usepackage{amsmath,amsfonts,bm}

\def\eqref#1{equation~\ref{#1}}

\def\1{\bm{1}}

\DeclareMathAlphabet{\mathsfit}{\encodingdefault}{\sfdefault}{m}{sl}
\SetMathAlphabet{\mathsfit}{bold}{\encodingdefault}{\sfdefault}{bx}{n}



%% file: sec/0_abstract.tex
\begin{figure*}[ht]
    \centering
    \includegraphics[width=0.95\textwidth]{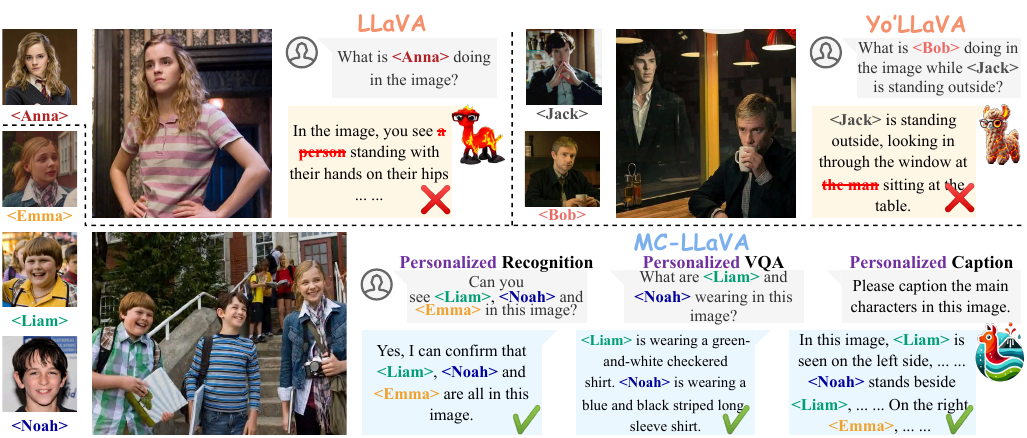}
    \caption{\textbf{LLaVA fails to understand user-provided concepts.} Existing methods like Yo'LLaVA mainly focus on single-concept personalization and can not generate accurate, personalized responses about multi-concepts. The proposed MC-LLaVA learns multiple concepts and can perform accurately in multi-concept scenario across various tasks. More qualitative comparisons are shown in Fig.~\ref{fig:additional_com}.}
    \label{fig:fig1}
\end{figure*}

\begin{abstract}
Current vision-language models (VLMs) show exceptional abilities across diverse tasks, such as visual question answering. To enhance user experience, recent studies have investigated VLM personalization to understand user-provided concepts. However, they mainly focus on single concepts, neglecting the existence and interplay of multiple concepts, which limits real-world applicability. This paper proposes MC-LLaVA, a multi-concept personalization paradigm. Specifically, MC-LLaVA employs a multi-concept instruction tuning strategy, effectively integrating multiple concepts in a single training step. To reduce the training costs, we propose a personalized textual prompt that uses visual token information to initialize concept tokens. Additionally, we introduce a personalized visual prompt during inference, aggregating location maps for enhanced recognition and grounding capabilities. To further push the performance upper bound, we incorporate an optional auxiliary loss, better enhancing the proposed personalized prompts. To decorate the VLM personalization research, we contribute a high-quality dataset. We carefully collect images with multiple characters and objects from movies and manually create question-answer samples for multi-concept scenarios, featuring superior diversity. Comprehensive experiments demonstrate that MC-LLaVA achieves impressive multi-concept personalized responses, paving the way for VLMs to become better user assistants. 
The code and dataset will be released at \href{https://github.com/arctanxarc/MC-LLaVA}{https://github.com/arctanxarc/MC-LLaVA}.

\end{abstract}

%% file: sec/1_intro.tex
\vspace{-7mm}
\section{Introduction}

\label{sec:intro}

Over the past few years, large language models (LLMs)~\cite{achiam2023gpt,bai2023qwen,touvron2023llama} have made significant advancements, proving their effectiveness in various applications and transforming the way humans interact with machines. In line with this trend, many vision-language models (VLMs)~\cite{liu2024visual,li2023blip,bai2024survey} have been proposed to connect vision encoders with LLMs for various vision-language tasks~\cite{deng2024can,zhou2024mathscape,lin2024draw}. Despite their success, VLMs face challenges when personalized responses are required, such as answering questions based on user-provided concepts. For example, given images with a concept $\langle \text{Anna} \rangle$, VLMs fail to generate property sentences, as shown in Fig.~\ref{fig:fig1}. This limitation hinders the smooth integration of VLMs into our daily lives.

\begin{figure}
    \centering
    \vspace{-5mm} 
    \includegraphics[width=0.48\textwidth]{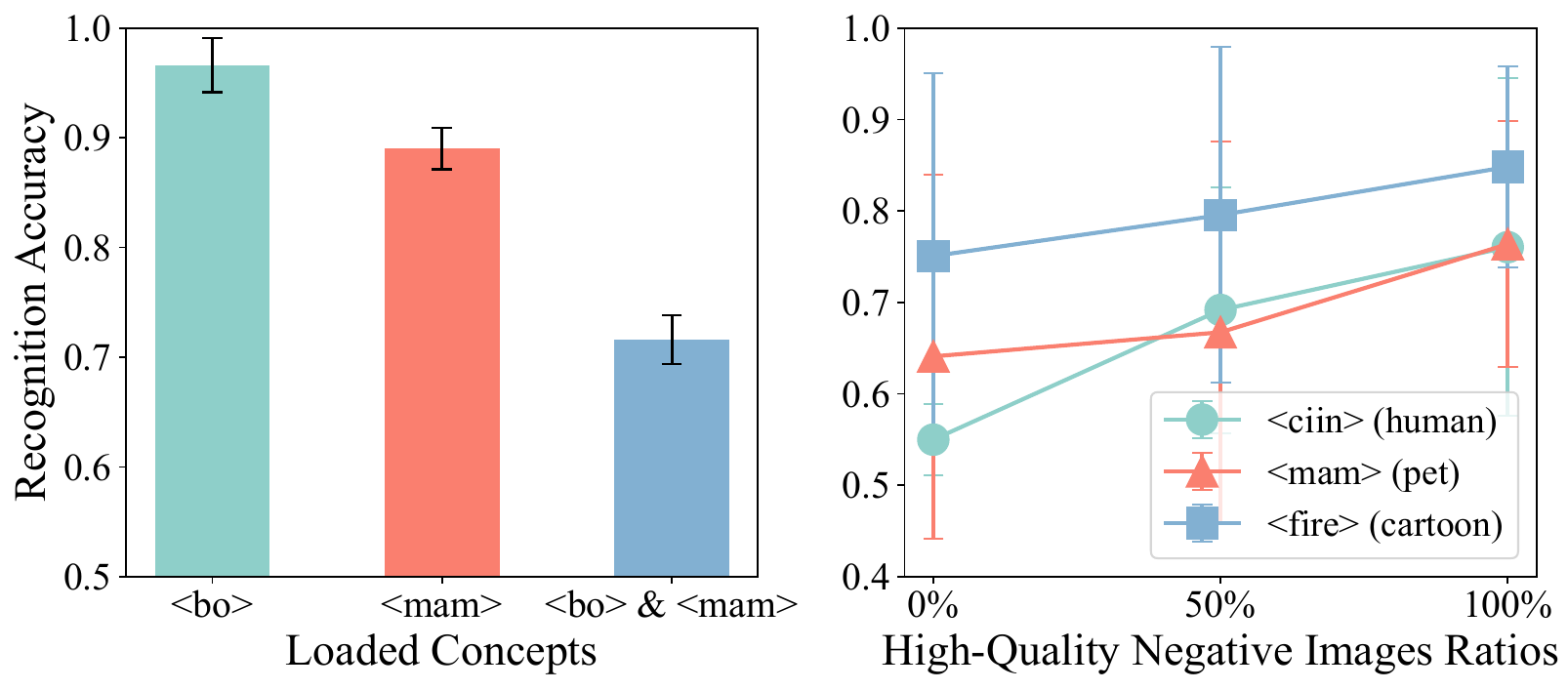}
    \caption{\textbf{Case studies of Yo’LLaVA.} The left panel shows the limitations of separate training, while the right emphasizes the significance of high-quality negative samples for its training process.}
    \label{fig:2_mot}
    \vspace{-7mm} 
\end{figure}

Although some methods~\cite{alaluf2025myvlm,pi2024personalized,nguyen2024yo} have produced impressive results in VLM personalization, they mainly concentrate on single-concept.
However, practical applications often require the integration of multiple concepts, a capability that is essential for large-scale deployment. Bridging this gap by personalizing multiple concepts—while preserving the model's inherent knowledge—is therefore critical for diverse downstream tasks.
Existing methods face significant scalability and precision bottlenecks in multi-concept scenarios. Tuning-based methods, such as Yo'LLaVA~\cite{nguyen2024yo}, train concepts separately, leading to performance degradation during parameter merging (Fig.~\ref{fig:2_mot} Left). Furthermore, personalizing concept requires extensive high-quality negative samples (Fig.~\ref{fig:2_mot} Right), significantly increasing data curation costs. While RAG-based~\cite{hao2024remember} and training-free approaches~\cite{seifi2025personalization} reduce the costs, they often fail in complex settings where visually similar subjects require precise disambiguation beyond simple retrieval. Consequently, robust multi-concept personalization remains a challenge, necessitating a systematic approach.

To solve the abovementioned problems, we introduce a novel method called MC-LLaVA, which ensures the accurate generation of personalized responses based on multiple concepts. MC-LLaVA considers multiple concepts together in a single training step rather than treating them independently. To reduce the cost of joint training, we pass all concept images through the VLM vision encoder and projection layer, using projected vision embeddings to initialize the concept tokens in personalized textual prompts. Our experiments show that this initialization can accelerate training and reduce dependence on high-quality negative samples. Additionally, MC-LLaVA enhances the model's perception capabilities by introducing a personalized visual prompt. We aggregate the location confidence maps based on concept tokens to create the personalized visual prompt. To further push the performance upper bound, we incorporate an optional mask-based supervision during joint training; by requiring the latent attention maps to localize concepts, the model achieves faster convergence and superior concept awareness. During inference, the integration of learnable tokens ensures better precise concept localization.

To advance research in multi-concept VLM personalization, datasets for training and testing are essential. Recent studies have developed datasets for personalized VLMs; however, these datasets focus only on single concepts. Furthermore, the types of questions and answers they address are limited to basic recognition and multiple-choice formats. The lack of datasets hinders the progress of multi-concept personalized VLMs. Therefore, we contribute a high-quality dataset by meticulously gathering images from concept-rich movies. We then utilize GPT-5~\cite{openai2025gpt5} to generate the initial question-answer samples and then manually refine the generated samples. Our dataset features diverse movie types and question-answer (QA) types. In total, our dataset includes approximately 2,000 images and 16,700 QA samples. To comprehensively evaluate multi-concept capability, we construct various tasks, including multi-concept recognition, visual grounding, QA, and captioning. Our dataset will facilitate future research in VLM personalization.

We summarize our contributions as follows:
\begin{itemize}[noitemsep, topsep=0pt, leftmargin=*]
\item We introduce MC-LLaVA, the first method designed for multi-concept VLM personalization, which employ personalized textual and visual prompts to learn various concepts and generate tailored responses effectively, reducing reliance on high-quality negative samples.
\item Queried GPT-5 approximately 100K times, we contribute a large-scale and high-quality dataset for training and testing multi-concept personalized VLMs.
\item We perform experiments on our dataset and two additional benchmarks, achieving state-of-the-art results among various tasks in single- and multi-concept personalization.
\end{itemize}

%% file: sec/2_related_work.tex
\label{sec:formatting}
\begin{figure*}[t]
    \centering
\includegraphics[width=0.95\textwidth]{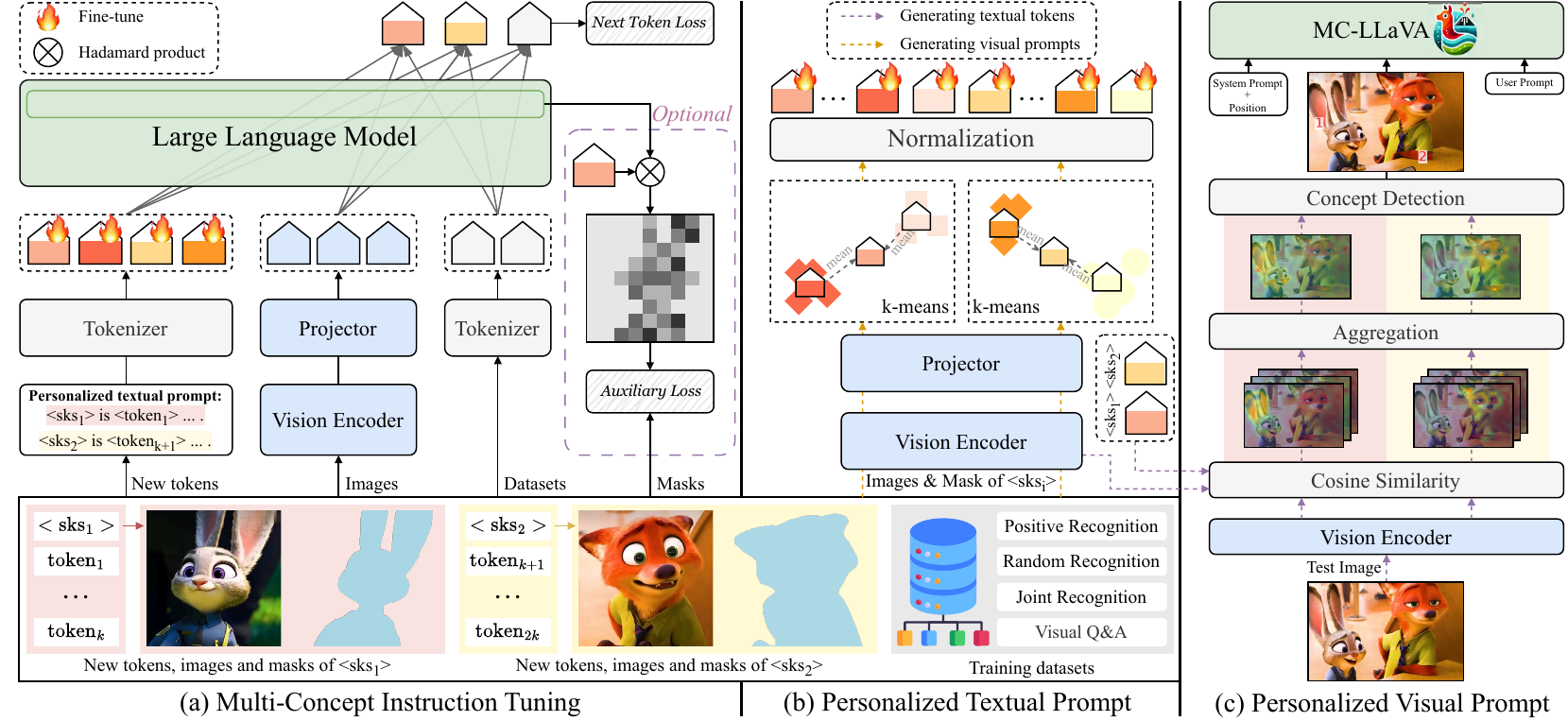}
    \caption{\textbf{Overview of MC-LLaVA.} (a) We use a multi-concept joint training strategy to learn the personalized textual prompts and classifier weights. (b) Given $m$ concepts, we utilize visual tokens obtained from K-means centroids to initialize the $m \times (k+1)$ concept tokens in personalized textual prompts, reducing costs associated with training. (c) During inference, we introduce personalized visual prompts by aggregating location confidence maps based on learned concept tokens. }
    \label{fig:fig2}
\end{figure*}

%% file: sec/3_method.tex
\section{Method}

Our method's pipeline is shown in Fig.~\ref{fig:fig2}. 
First, we propose multi-concept instruction tuning for MC-LLaVA (Sec.~\ref{sec:mc_tune}). 
Next, we introduce a personalized textual prompt for multi-concept and optional loss to accelerate convergence and enhance the feature representation for downstream modules (Sec.~\ref{sec:textualprompt}). 
Finally, we propose a personalized visual prompt to boost recognition and grounding (Sec.~\ref{sec:visualprompt}).

\subsection{Multi-Concept Instruction Tuning}\label{sec:mc_tune}

Given a pre-trained VLM and multiple user-provided concepts, our goal is to introduce new concepts by expanding the vocabulary and learning the personalized textual prompts, while preserving the model’s knowledge. 
Instead of training concepts separately, as in Yo'LLaVA~\cite{nguyen2024yo}, we propose a joint training approach that considers multiple concepts together and simultaneously learns personalized textual prompts and a classifier (see Fig.~\ref{fig:fig2}(a)). 
Specifically, for $m$ concepts $\{C^j\}_{j=1}^m$, each with $n$ images $\{I^i\}_{i=1}^n$, we define $k+1$ learnable tokens per concept:
\begin{equation}
    \bigcup_{j=1}^{m} \{\langle\text{sks}_j\rangle, \  \langle\text{token}_{(j-1)k+1}\rangle \ldots \langle\text{token}_{jk}\rangle\}
\label{eq:ptp}
\end{equation}
where each $\langle \text{sks}_j \rangle$ serves as a unique concept identifier. As these concepts are new to the VLM, we expand the vocabulary by adjusting the language model classifier's weights $W$ from $D \times N$ to $D \times (N+m)$, where $D$ is the feature dimension and $N$ is the original vocabulary size.

To train MC-LLaVA, we construct training samples in the form $(I, X_q, X_a)$, where $I$ is the input image, $X_q$ is the question, and $X_a$ is the answer. Following prior works such as Yo'LLaVA~\cite{nguyen2024yo}, we adopt standard dialogue formats (e.g., positive recognition, random recognition, and conversation tasks) in our training pipeline. 
Additionally, we design a novel joint recognition task. In our joint training, we pair text and image samples corresponding to different concepts within the same scenario, to generate cross-concept pairs. 
This inter-concept negative sampling strategy produces $m\times(m-1)\times n$ negative samples for scenarios containing $m$ concepts with $n$ images.

Details on training set construction are provided in Sec.~\ref{sec:4}. All mentioned data examples are shown in the Appendix. In our joint training framework, parameters to be updated are:
\begin{equation}
\theta = \{ \langle \text{sks}_{1:m} \rangle, \langle \text{token}_{1:mk} \rangle, W(:, N+1 : N + m) \}
\end{equation}
We utilize standard masked language modeling loss to compute probability of target responses $X_a$:
\begin{equation}
\mathcal{L}_{LM}(X_a | I, X_q, \theta) = -\sum_{t=1}^{T} \log P(X_{a,t} | I, X_q, X_{a,<t}, \theta)
\end{equation}
where \( T \) is the length of the answer \( X_a \), \( X_{a,t} \) denotes the \( t \)-th word in the answer, and \( P(X_{a,t} | I, X_q, X_{a,<t}, \theta) \) represents the probability of predicting the \( t \)-th word given the input image \( I \), the question \( X_q \), all preceding words in the answer \( X_{a,<t} \), and the parameters \( \theta \).

\subsection{Personalized Textual Prompt}
\label{sec:textualprompt}


To train the newly introduced tokens, we construct the personalized system prompt by sequentially appending prompts for each concept in the given scenario. 
Specifically, for each concept $C^j$, we add the prompt: ``$\langle \text{sks}_j \rangle$ is $\langle \text{token}_{(j-1)k+1} \rangle \dots \langle \text{token}_{jk} \rangle$''.

To decrease reliance on high-quality negative samples, which are challenging to acquire, we employ a visual-based token initialization strategy, as illustrated in Fig.~\ref{fig:fig2}(b). Instead of random initialization, we leverage Grounded-SAM~\cite{ren2024grounded} with the prompt ``the main character'' to extract foreground masks from the training images. 
Subsequently, by applying these masks to the visual features of the training images, we extract a set of features that capture the fine-grained semantic granularity of the specific concept. 
We then apply k-means clustering to these refined features to obtain $k$ centroids, which are used to initialize the $k$ learnable tokens. This approach effectively accelerates convergence, as demonstrated in Sec.~\ref{exp:epoch}.



To further ensure that the learnable concept tokens specifically attend to the relevant visual regions, we introduce an optional attention-guided visual alignment objective (see Fig.~\ref{fig:fig2}(a)). 
This auxiliary supervision grounds the concept tokens to the correct visual entity, reducing hallucinations in multi-concept scenarios.
We extract the attention weights $\mathbf{A}^{(l)}$ from the last $K$ layers. 
Let $\mathcal{I}_{c}$ and $\mathcal{I}_{v}$ denote the indices of the concept token ($\langle \text{sks} \rangle$) and visual tokens, respectively. We compute a spatial attention map $M_{attn} \in \mathbb{R}^{|\mathcal{I}_{v}|}$ by averaging the weights across layers, heads ($H$), and $\mathcal{I}_{c}$:
\begin{equation}
    M_{attn}[j] = \frac{1}{K \cdot H \cdot |\mathcal{I}_{c}|} \sum_{l, h, i \in \mathcal{I}_{c}} \mathbf{A}^{(l)}_{h, i, j}, \quad \forall j \in \mathcal{I}_{v}
\end{equation}
We treat $M_{attn}$ as a predicted soft mask and align it with the ground truth binary mask $M_{gt}$ (provided by Grounded-SAM) by minimizing the auxiliary loss:
\begin{equation}
    \mathcal{L}_{attn} = 1 - \frac{2 \sum_{j} (M_{attn}[j] \cdot M_{gt}[j]) + \epsilon}{\sum_{j} M_{attn}[j] + \sum_{j} M_{gt}[j] + \epsilon}
\end{equation}
where $\epsilon$ is a smoothing term. The final training objective is a weighted sum of the standard next-token prediction loss and the auxiliary loss: $\mathcal{L} = \mathcal{L}_{LM} + \lambda \mathcal{L}_{attn}$.

\begin{figure*}[!t]
    \centering
    \includegraphics[width=0.95\textwidth]{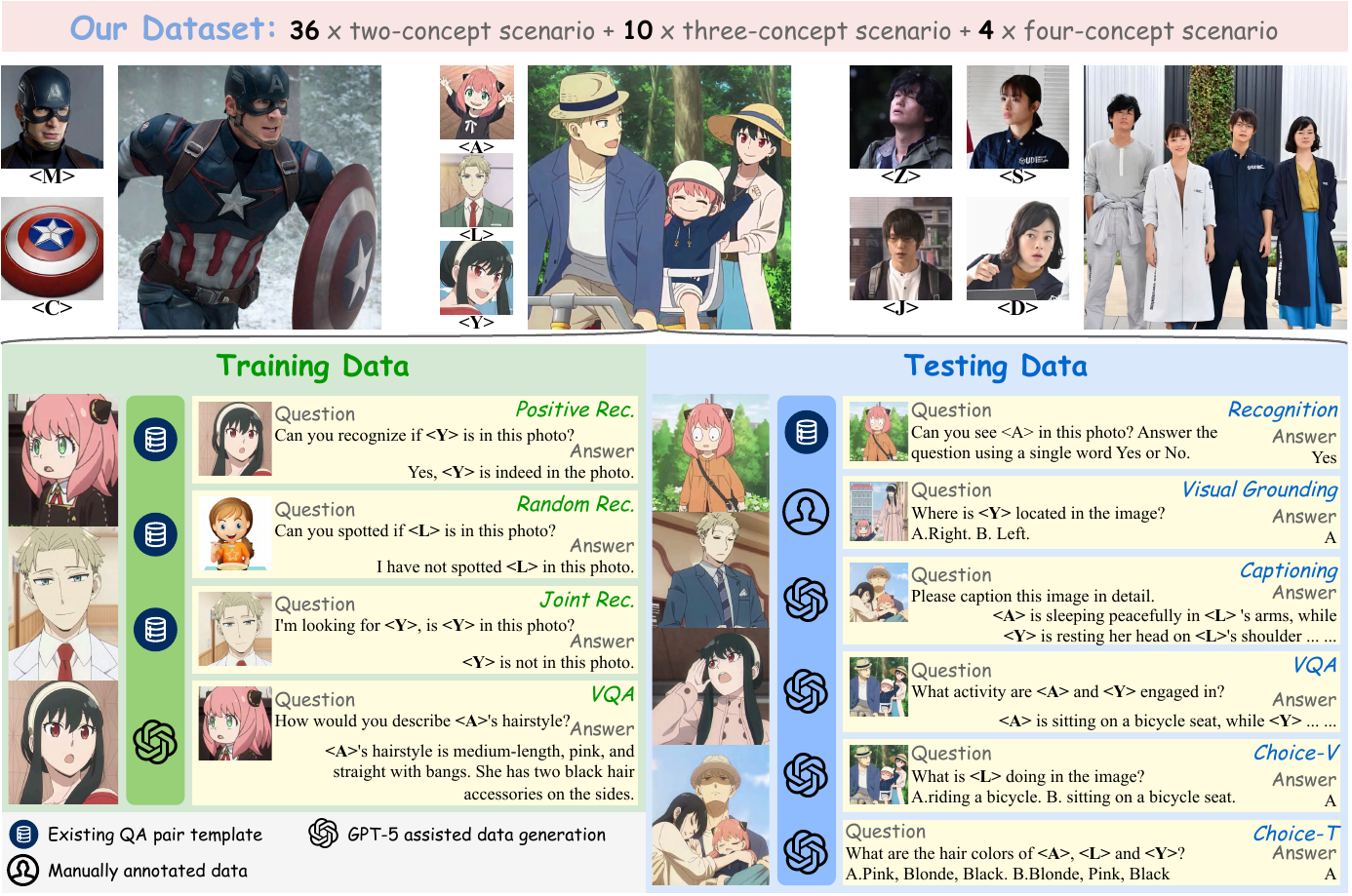}
    \caption{\textbf{Examples of the proposed multiple concept personalization dataset.}  The dataset includes not only adults but also children, animals and objects, derived from cartoons and movies. To facilitate visualization, concept identifiers have been abbreviated using letters.}
    \label{fig:dataset}
    \vspace{-4mm}
\end{figure*}

\subsection{Personalized Visual Prompt}
\label{sec:visualprompt}


While textual tokens capture semantic attributes, relying solely on them may be inadequate for precise spatial grounding in multi-concept scenarios. 
To address this, we propose a Personalized Visual Prompt mechanism based on Set-of-Mark (SOM)~\cite{yang2023set}. 
As shown in Fig.~\ref{fig:fig2}(c), this module generates explicit spatial cues by computing a location confidence map, leveraging both the preserved training features and the learned concept tokens (optional).

Given a test image $I_t$, we extract its spatial visual tokens $\mathbf{V}_t \in \mathbb{R}^{N \times D}$ using the visual encoder, where $N$ is the number of patches. To robustly localize the $j$-th concept $C^j$, we employ a hybrid similarity metric that combines retrieval-based matching with token-based guidance.

First, we compute a reference similarity map $M_{ref}$ based on the support set of training images $\{I_k\}_{k=1}^{L_j}$ stored for concept $C^j$. 
For each spatial position $p$, $M_{ref}[p]$ is the average cosine similarity between $\mathbf{V}_t$ and the support features. Second, to exploit the visual-textual alignment achieved in Sec.~\ref{sec:textualprompt}, we utilize the learned token embedding $\mathbf{e}_j$ of $\langle \text{sks}_j \rangle$. We compute a token-guided map $M_{token}$ as:
\begin{equation}
    M_{token}[p] = \frac{\mathbf{V}_t[p] \cdot \mathbf{e}_j^\top}{\|\mathbf{V}_t[p]\| \cdot \|\mathbf{e}_j\|}
\end{equation}
The final localization map $M_{final}$ is a weighted fusion of these two components:
\begin{equation}
    M_{final} = \beta M_{token} + (1 - \beta) M_{ref}
\end{equation}

To construct the visual prompt, we normalize $M_{final}$ and apply a confidence threshold $\tau$ to filter out background noise. The representative coordinate is identified as the location with the maximum response. If the activated region implies the concept's presence, we append a spatial indicator to the system prompt: ``$\langle\text{sks}_j\rangle \text{ is located at Mark } j$'', enhancing the model's grounding capability within inference.

%% file: sec/4_dataset.tex
\section{Multi-Concept Instruction Dataset}
\label{sec:4}

In this section, we provide a comprehensive explanation of the process involved in creating our multi-concept instruction dataset. The dataset includes a training set with single-concept images and a testing set containing both single- and multi-concept images, totaling approximately 2K images. 
Tab.~\ref{tab:combined} compares our dataset with recent datasets. Our dataset is superior due to its support for multiple concepts, more advanced captions, and a larger sample size.

\begin{table}[b]
\centering
\caption{Statistics and comparison of our dataset. (Top) Comparison with representative datasets. (Bottom) Detailed statistics of our dataset across different concept counts.}
\label{tab:dataset_all}
\setlength{\tabcolsep}{1.2mm} 
\renewcommand{\arraystretch}{1.3} 
\footnotesize 

\begin{tabular}{c|c|c|c|c}
\hline
\multicolumn{5}{c}{\textbf{Dataset Comparison}} \\ \hline
\textbf{Dataset} & \textbf{Concept} & \textbf{Caption} & \multicolumn{2}{c}{\textbf{Samples}} \\ \hline
MyVLM & Single & Human & \multicolumn{2}{c}{0.3K} \\
Yo'LLaVA & Single & Human & \multicolumn{2}{c}{0.6K} \\ \hline
\rowcolor[gray]{0.9} \textbf{Ours} & \makecell[c]{\textbf{Single \&} \\ \textbf{Multi}} & \makecell[c]{\textbf{GPT-4o \&} \\ \textbf{Human}} & \multicolumn{2}{c}{\textbf{2.0K}} \\ \hline
\multicolumn{5}{c}{} \\[-0.8em] 
\hline
\multicolumn{5}{c}{\textbf{Detailed Statistics (Ours)}} \\ \hline
\makecell[c]{\textbf{Total} \\ \textbf{Scenarios}} & \textbf{Concept} & \textbf{Scenario} & \textbf{Samples} & \textbf{QA Pairs} \\ \hline
\multirow{3}{*}{\textbf{50}} & 2 & 36 & 1,260 & 9,900 \\ \cline{2-5}
 & 3 & 10 & 500 & 4,350 \\ \cline{2-5}
 & 4 & 4 & 260 & 2,444 \\ \hline
\end{tabular}
\label{tab:combined}
\end{table}

\subsection{Image Data Collection}
\label{sec:4.1}

The field of VLM personalization lacks large-scale, high-quality datasets. Existing datasets rely mainly on manually captured photos, which are challenging to obtain in multi-concept scenarios. Moreover, privacy concerns hinder the scalability of such data collection. 
To overcome these limitations, we systematically curate images from a diverse selection of animated and live-action films worldwide, ensuring broad cultural and artistic coverage. This approach facilitates the collection of multi-concept data, focusing on instances where multiple concepts co-occur. To prevent the model from relying on pre-trained knowledge for concept recognition, we reassign each concept a generic label (e.g., $\langle \text{Anna} \rangle$). Our dataset encompasses a wide range of concepts, including animals, human characters, and objects, providing a rich and diverse resource for model training. 

For training, we collect ten images per concept, ensuring distinct visual characteristics with diverse appearances, contexts, and backgrounds to enhance generalization. The test set includes both single- and multi-concept images. Single-concept images follow the same collection strategy as the training set. For multi-concept images, we define specific pairs and select frames where all concepts are clearly visible, maintaining a balanced distribution.  
To ensure fairness and high data quality, the data collection process was collaboratively designed by a team of ten members, comprising university students and researchers from technology companies, minimizing subjectivity and bias. The upper section of Fig.~\ref{fig:dataset} presents sampled images from the dataset.

\subsection{GPT-5-Assisted Data Generation}
\label{sec:4.2}

After acquiring the training and testing images, we employ GPT-5 to generate question-answer pairs. For the training images, we first prompt GPT-5 to generate a diverse set of general questions related to each concept. Subsequently, we manually select ten questions per concept that provide broad coverage, ensuring the model can effectively learn the concept. These selected images and questions are then input into GPT-5 to generate more refined answers.

For the testing images, we utilize GPT-5 to create VQA dialogues and multiple-choice questions that prioritize the visual content of the images rather than the broader concept-related questions used in the training set. Specifically, if a test image contains only a single concept, the questions focus solely on that concept. Conversely, for images containing multiple concepts, the questions include both single-concept and multi-concept queries to ensure a more realistic evaluation. The generated questions are manually curated to maintain quality, after which images and prompts are fed into GPT-5 to obtain answers, which serve as ground truth. To construct our dataset, we queried GPT-5 approximately 100K times. The Fig.~\ref{fig:dataset} presents examples of curated data.

%% file: sec/5_exp.tex
\section{Experiment}


\subsection{Experimental Setup}
\label{sec:5.1}
\noindent \textbf{Evaluation Datasets.}
In addition to conducting experiments on our dataset, we further validate our method on two established benchmarks: Yo’LLaVA~\cite{nguyen2024yo} and MyVLM~\cite{alaluf2025myvlm}. The Yo’LLaVA dataset encompasses 40 diverse concepts (e.g., objects, buildings, people), with each concept represented by 4 to 10 images. In comparison, MyVLM focuses on 29 object-centric concepts, providing a minimum of 10 images for each. All data from these two benchmarks are sourced from daily lifes.

\noindent \textbf{Baselines.}
MC-LLaVA is compared with naive prompting and other VLM personalization methods:
\begin{itemize}[noitemsep, topsep=0pt, leftmargin=*]
    \item \textbf{Yo'LLaVA}~\cite{nguyen2024yo}: We adopt two settings, namely Yo’LLaVA-S and -M. Both settings train each concept separately—Yo'LLaVA-S loads parameters for one concept (supporting only single-concept queries), while Yo'LLaVA-M fuses these tokens with extended classification head parameters to enable multi-concept queries.
    \item \textbf{RAP-MLLM}~\cite{hao2024rememberretrievegenerateunderstanding}: We utilize the RAP-LLaVA model and follow the RAP-MLLM approach to construct a personalized database for each dataset.
    \item \textbf{PeKit}~\cite{seifi2025personalization}: Due to the lack of publicly available source code, we re-implemented the framework following its original architecture. PeKit functions by encoding patch-level features from reference images into a persistent memory module $\mathcal{M}$, which are then dynamically retrieved via a retrieval module $\mathcal{R}$ to serve as visual prompts for personalized inference. 
\end{itemize}
Details of other compared baselines can be found in the Appendix. All baselines and our method are evaluated on three datasets. For our dataset, we report results for single- and multi-concept questions. In the subsequent results, all outcomes are averaged over three runs with different seeds.

\noindent \textbf{Implementation Details.} For training, we use $10$ images per concept and set the number of concept tokens ($k$) to $16$. We fine-tune LLaVA-1.5-13B with the AdamW~\cite{kingma2014adam} optimizer, employing a learning rate of $0.001$ over $15$ epochs. More details are provided in the Appendix.

\begin{table}
    \centering
    \scriptsize
    \setlength{\tabcolsep}{0.6mm}  
    \renewcommand{\arraystretch}{0.6}  
    \caption{\textbf{Comparison of Rec. and VG capabilities.} P = Prompt; Rec. = Recognition; VG = Visual grounding. AL = Auxiliary loss. The \colorbox{red!15}{best} and \colorbox{blue!15}{second best} are highlighted.}
    \label{exp:rec}
    \begin{tabular}{c|c||ccc|c|c|c}
    \toprule
    \multicolumn{2}{c||}{\textbf{Evaluation Dataset}} & \multicolumn{4}{c|}{\textbf{MC-LLaVA}} & \textbf{Yo'LLaVA} & \textbf{MyVLM} \\
    \midrule
    \multirow{2}{*}{\textbf{Method}} & \multirow{2}{*}{\textbf{Tokens}} & \multicolumn{3}{c|}{\textbf{Rec.}} & \multirow{2}{*}{\textbf{VG}} & \textbf{Rec.} & \textbf{Rec.} \\
    & & Single & Multi & Weight & & Single & Single \\

    \midrule
    \textcolor{gray}{\textbf{GPT4o+P}} & \textcolor{gray}{$10^1$} & \textcolor{gray}{0.746} & \textcolor{gray}{0.822} & \textcolor{gray}{0.781} & \textcolor{gray}{0.699} & \textcolor{gray}{0.856} & \textcolor{gray}{0.891} \\
    \midrule
    \textbf{LLaVA} & $0$ & 0.500 & 0.501 & 0.500 & 0.458 & 0.500 & 0.500 \\
    \rowcolor[gray]{0.95} \textbf{LLaVA+P} & $10^1$ & 0.594 & 0.549 & 0.573 & 0.528 & 0.819 & 0.732 \\
    \textbf{LLaVA+P} & $10^2$ & 0.590 & 0.590 & 0.590 & 0.567 & 0.650 & 0.674 \\
    \rowcolor[gray]{0.95}   \textbf{MyVLM} & $1$ & 0.795 & - & 0.795 & 0.688 & 0.911 & 0.938 \\
    \textbf{Yo'LLaVA-S} & $10^1$ &  0.841 & - & 0.841 & 0.702 &   0.924 &  0.964 \\
    \rowcolor[gray]{0.95} \textbf{Yo'LLaVA-M} & $10^1$ & 0.744 & 0.729 & 0.737 & 0.612 &   0.924 &  0.964 \\
    \textbf{RAP-MLLM} & $10^2$ & 0.747 & 0.688 & 0.713 & 0.719 &0.845 & 0.870 \\
    \rowcolor[gray]{0.95}\textbf{PeKit} & $0$ & 0.791 & 0.704 & 0.754 & \cellcolor{blue!15}0.723  & 0.946 & 0.972 \\
    \textbf{Ours} & $10^1$ & \cellcolor{blue!15}0.912 & \cellcolor{blue!15}0.845 & \cellcolor{blue!15}0.878 & \cellcolor{blue!15}0.723 & \cellcolor{blue!15}0.947 & \cellcolor{blue!15}0.975 \\
    \rowcolor[gray]{0.95} \textbf{Ours + AL} & $10^1$ & \cellcolor{red!15}0.932 & \cellcolor{red!15}0.867 & \cellcolor{red!15}0.899 & \cellcolor{red!15}0.748 & \cellcolor{red!15}0.962 & \cellcolor{red!15}0.984\\
    \bottomrule
    \end{tabular}
    \vspace{-4mm} 
\end{table} 

\subsection{Recognition and Visual Grounding Ability}
\label{sec:5.2}

\begin{table*}[!t]
\centering
\caption{\textbf{Comparison of the question-answering capabilities.} P = Prompt. T = Tokens. Yo'L = Yo'LLaVA. RAP = RAP-MLLM. AL = Auxiliary loss. The \colorbox{red!15}{best} and \colorbox{blue!15}{second best} performances are highlighted.}
\scriptsize
    \setlength{\tabcolsep}{0.4mm}  
    \renewcommand{\arraystretch}{0.7}   
\resizebox{\textwidth}{!}{
\begin{tabular}{c|c||ccc|ccc|ccc|ccc|c|c}
\toprule
\multicolumn{2}{c||}{\textbf{\makecell{Evaluation \\ Dataset}}} &  \multicolumn{12}{c|}{\textbf{MC-LLaVA}} & \textbf{Yo'L} & \textbf{MyVLM} \\
\midrule
\multirow{2}{*}{\textbf{Method}} & \multirow{2}{*}{\textbf{T}} & \multicolumn{3}{c|}{\textbf{Choice-V Acc.}} & \multicolumn{3}{c|}{\textbf{Choice-T Acc.}} & \multicolumn{3}{c|}{\textbf{VQA BLEU}} & \multicolumn{3}{c|}{\textbf{Caption Recall}} & \textbf{\makecell{Choice-V \\ \&T Acc.}} & \textbf{\makecell{Caption \\ Recall}} \\
& & Single & Multi  & Weight  & Single & Multi  & Weight  & Single & Multi  & Weight  & Single & Multi  & Weight  & Single & Single \\
\midrule
\textcolor{gray}{\textbf{GPT4o+P}} & \textcolor{gray}{$10^1$} & \textcolor{gray}{0.888} & \textcolor{gray}{0.889} & \textcolor{gray}{0.889} & \textcolor{gray}{0.712} & \textcolor{gray}{0.680} & \textcolor{gray}{0.702} & \textcolor{gray}{0.728} & \textcolor{gray}{0.651} & \textcolor{gray}{0.701} & \textcolor{gray}{0.836} & \textcolor{gray}{0.816} & \textcolor{gray}{0.830} &  \textcolor{gray}{0.840} & \textcolor{gray}{0.969} \\
\midrule
\textbf{LLaVA} & $0$ & 0.806 & 0.802 & 0.804 & 0.411 & 0.264 & 0.353 & 0.317 & 0.208 & 0.280 & 0.096 & 0.050 & 0.082 & 0.721  & 0.021 \\
\rowcolor[gray]{0.95} \textbf{LLaVA+P} & $10^1$ & 0.837 & 0.781 & 0.817 & 0.597 & 0.535 & 0.553 & 0.428 & 0.364 & 0.407 & 0.108 & 0.160 & 0.123 & 0.835 & 0.207 \\
\textbf{LLaVA+P} & $10^2$ &  0.841 & 0.785 & 0.825 & 0.646 & 0.630 & 0.635 & 0.436 & 0.375 & 0.415 & 0.054 & 0.122 & 0.075 & 0.728 & 0.211 \\
\rowcolor[gray]{0.95} \textbf{MyVLM} & $1$ & 0.779 & - & 0.779 & - & - & - & 0.640 & - & 0.640 & 0.714 & - & 0.714 & 0.845  & 0.921 \\
\textbf{Yo'L-S} & $10^1$ & 0.801 & - & 0.801 & 0.703 & - & 0.703 & 0.643 & - & 0.643 & 0.701 & - & 0.701 & 0.896  & 0.931 \\
\rowcolor[gray]{0.95} \textbf{Yo'L-M} & $10^1$ & 0.688 & 0.602 & 0.655 & 0.684 & 0.594 & 0.658 & 0.604 & 0.557 & 0.588 & 0.622 & 0.611 & 0.619 &  0.896  & 0.931 \\
\textbf{RAP} & $10^2$ & 0.832 & 0.690 & 0.784 & 0.709 & 0.656 & 0.685 & 0.524 &0.523 & 0.524 & 0.711 & 0.748 & 0.723 & 0.917  & 0.937 \\
\rowcolor[gray]{0.95} \textbf{PeKit} & $0$ & 0.874  & 0.867 & 0.872 & 0.712 & 0.657 & 0.692 & 0.635 & 0.601 & 0.623 & 0.729 & 0.743 & 0.733 & 0.922 &  0.970 \\
\textbf{Ours} & $10^1$ & \cellcolor{blue!15}0.877 & \cellcolor{blue!15}0.905 & \cellcolor{blue!15}0.890 & \cellcolor{blue!15}0.723 & \cellcolor{blue!15}0.695 & \cellcolor{blue!15}0.709 & \cellcolor{blue!15}0.679 & \cellcolor{blue!15}0.611 & \cellcolor{blue!15}0.658 & \cellcolor{blue!15}0.741 & \cellcolor{blue!15}0.763 & \cellcolor{blue!15}0.754 & \cellcolor{blue!15}0.925 & \cellcolor{blue!15}0.972 \\
\rowcolor[gray]{0.95} \textbf{Ours+AL} & $10^1$ & \cellcolor{red!15}0.909 & \cellcolor{red!15}0.920 & \cellcolor{red!15}0.914 & \cellcolor{red!15}0.729 & \cellcolor{red!15}0.698 & \cellcolor{red!15}0.718 & \cellcolor{red!15}0.701 & \cellcolor{red!15}0.638 & \cellcolor{red!15}0.681 & \cellcolor{red!15}0.772 & \cellcolor{red!15}0.784 & \cellcolor{red!15}0.775 & \cellcolor{red!15}0.942 & \cellcolor{red!15}0.975 \\
\bottomrule
\end{tabular}
}
\label{exp:vqa}
\vspace{-5mm}
\end{table*}

To evaluate the model's recognition ability, we conduct experiments on the MC-LLaVA, Yo'LLaVA, and MyVLM datasets. For the latter two, we adhere to the evaluation protocols specified in their respective papers. 
For our dataset, images containing the queried concept are treated as positive examples, whereas negative examples are selected from images depicting other concepts in the same scenario as well as from unrelated scenarios. 
Details are provided in the Appendix. Each test uses the query: ``Can you see {$\langle \text{sks}_i \rangle$} in this photo? Answer with a single word: Yes or No.''

Test data is categorized as single-concept or multi-concept based on the number of concepts queried in each question. 
To mitigate potential sample imbalance, 
we follow Yo'LLaVA~\cite{nguyen2024yo} and report arithmetic mean of the yes and no recall for positive and negative samples.


As summarized in Tab.~\ref{exp:rec},
Vanilla LLaVA achieves the lowest scores, because it lacks any additional concept information. 
Moreover, simply adding personalized prompts (LLaVA+P) results in only marginal improvements, suggesting that only a textual prompt does not effectively personalize LLaVA. 
Yo'LLaVA-M, without leveraging visual features, exhibits reduced performance on both single- and multi-concept queries, possibly due to confusion between different concepts. 
RAP-MLLM employs extra recognition modules and supports multi-concept queries by using top-K selection mechanism, however, this approach may occasionally struggle to accurately detect when a concept is absent. 
In comparison, our proposed MC-LLaVA method  uses fewer tokens and achieves state-of-the-art (SOTA) recognition performance in both single- and multi-concept scenarios. Notably, MC-LLaVA outperforms GPT4o+P, demonstrating its integrated textual and visual prompt design delivers richer concept-specific information than GPT-4o.

To further assess the model's visual grounding capability, particularly in multi-concept scenarios, we manually annotate the locations of each concept in multi-concept images. The model is then tested using a multiple-choice format to determine each concept’s position: ``Where is {$\langle \text{sks}_i \rangle$} located in this photo? A. Left. B. Middle. C. Right.'' The results are reported in terms of accuracy.
Because the visual grounding task further evaluates the model’s ability to localize concepts in multi-concept scenarios, most models exhibit a slight performance drop compared to the recognition task. Notably, RAP-MLLM secures second-best, likely due to its pre-training data  incorporating grounding-related tasks that enhance localization capabilities, whereas MC-LLaVA, with personalized visual prompt, achieves SOTA performance.

\subsection{Question Answering and Captioning Ability}
\label{sec:5.3}

We evaluate QA performance on two datasets: our proposed and Yo'LLaVA dataset. For Yo'LLaVA, we utilize their publicly available 571 multiple-choice QA test samples. Our dataset evaluation employs two complementary approaches: (1) multiple-choice QA covering both visual and text-based questions (2) open-ended visual question answering (VQA).

For visual questions, 
we construct multiple-choice questions across different concept configurations, including single- and multi-concept questions for composite scenarios. The dataset contains 1,180 single- and 600 multi-concept choice questions.
To mitigate the influence of random guessing in multiple-choice questions, we create corresponding VQA pairs (equal in number to the multiple-choice questions) for comprehensive evaluation. We employ two evaluation metrics: accuracy for choice selection and BLEU~\cite{papineni2002bleu} for text in open-ended responses.

As shown in Tab.~\ref{exp:vqa}, MC-LLaVA achieves significantly improved performance in VQA, with results comparable to GPT-4o. 
In the multiple-choice QA evaluation, MC-LLaVA delivers competitive performance with GPT-4o and outperforms all other baselines.
In the open-ended VQA setting, our method attains an overall BLEU score of 0.658, ranking second only to GPT-4o. 
Notably, RAP-MLLM, which is pre-trained on large scale data, tends to generate shorter responses and consequently scores the lowest in BLEU.


To evaluate whether the language model has truly memorized the new concepts, we designed 590 single-concept and 250 multi-concept text-only multiple-choice questions that focus on each concept's intrinsic characteristics. 
In the text-only QA task (Tab.~\ref{exp:vqa}), LLaVA+Prompt shows a notable performance boost as the number of prompt tokens increases, thanks to the enriched textual context. 
Among all models, MC-LLaVA achieves SOTA performance.

We conduct captioning evaluations on both our dataset and the MyVLM~\cite{alaluf2025myvlm} dataset, following the evaluation method proposed by MyVLM.  
The metric's detailed calculation method and the prompt are provided in the Appendix.
As shown in Tab.~\ref{exp:vqa}, on MC-LLaVA dataset, our method achieves a weighted captioning recall of 0.754—outperforming most baselines. On MyVLM dataset, MC-LLaVA nearly matches GPT4o and exceeds other models. 

\begin{figure*}[!t]
    \centering
\includegraphics[width=0.95\textwidth]{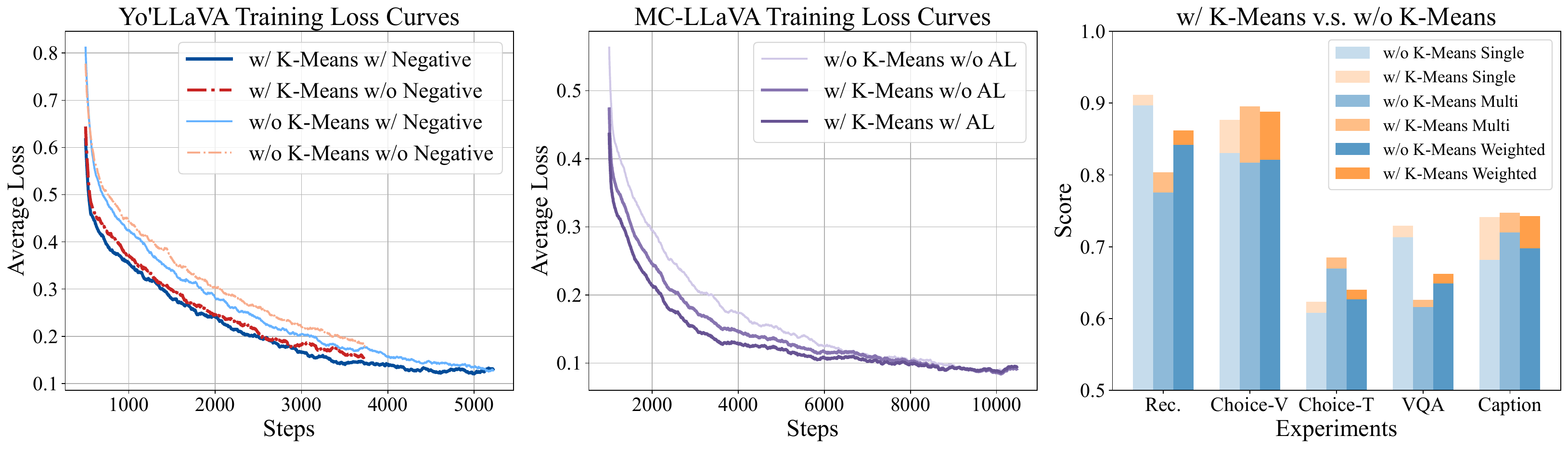}
    \vspace{-2mm}
    \caption{\textbf{Ablation study of the impact of initialization and auxiliary loss on training convergence and downstream performance.}}
    \label{fig:ablation_epoch}
    \vspace{-2mm}
\end{figure*}

\begin{figure*}[!t]
    \centering
\includegraphics[width=0.95\textwidth]{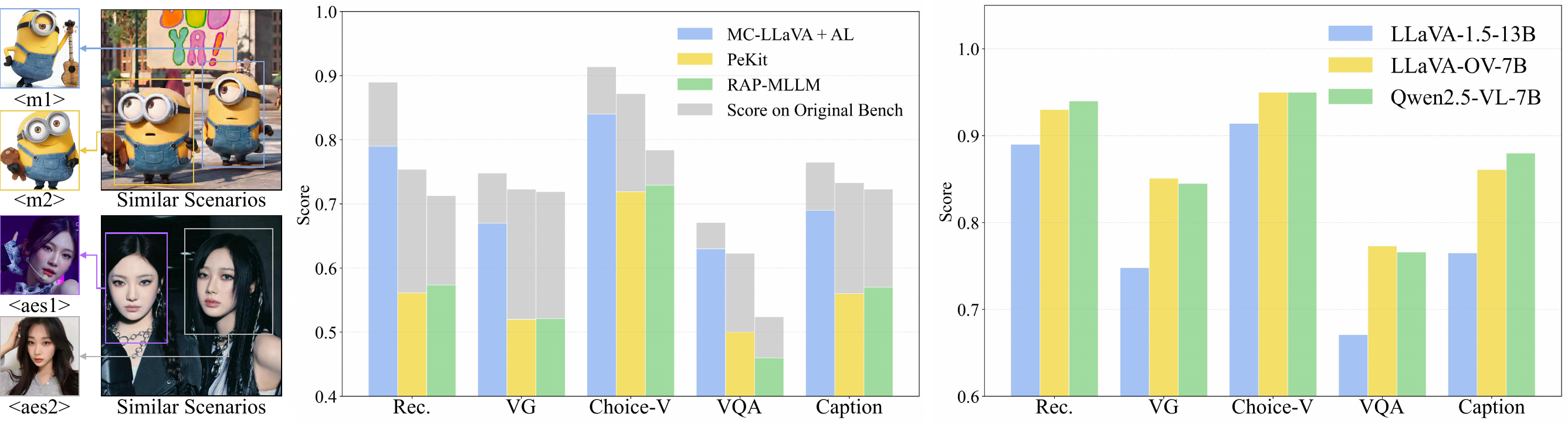}
    \vspace{-2mm}
    \caption{\textbf{Performance on visually similar scenarios (left) and analysis of generalizability across different VLM backbones (right)}}
    \label{fig:ablation2}
    \vspace{-4mm}
\end{figure*}

\subsection{Ablation and Analysis}
\label{sec:5.5}
\noindent \textbf{Training Progress.}
\label{exp:epoch}We conduct experiments on Yo'LLaVA and MC-LLaVA to evaluate the impact of negative samples, initialization, and auxiliary loss. As shown in the two leftmost images of Fig.~\ref{fig:ablation_epoch}, concept token initialization accelerates convergence, further boosted by the auxiliary loss. Notably, K-means initialization yields very similar loss curves regardless of negative samples, highlighting reduced dependency on high-quality negative data.

\noindent \textbf{Concept Token Initialization.}
We assess the effectiveness of MC-LLaVA for initializing concept tokens across downstream tasks.
As shown in the right image of Fig.~\ref{fig:ablation_epoch}, using the initialization leads to observed performance improvements across all tasks. 
In addition to significant improvements in various visual tasks, our design also results in a slight enhancement in text-only tasks. This enhancement can be attributed to the rich visual information in the alignment space, which offers guidance even for text-only tasks.

\begin{table}
\centering
\caption{\textbf{Ablations on modules.} HNS = High-quality negative samples. AL = Auxiliary loss. Init. = Initialization. P. = Prompt.}
\label{tab:module}
\setlength{\tabcolsep}{0.1mm}  
\scriptsize
\begin{tabular}{lccccc}
\toprule
 \textbf{Module/Task} & \multicolumn{1}{c}{\textbf{Rec.}} & \multicolumn{1}{c}{\textbf{VG}} & \multicolumn{1}{c}{\textbf{Choice-V}}  & \multicolumn{1}{c}{\textbf{VQA}} & \multicolumn{1}{c}{\textbf{Captioning}} \\
\midrule
\textbf{Baseline} & 0.737 & 0.612 & 0.655 & 0.588 & 0.619  \\
\rowcolor[gray]{0.95} \textbf{- HNS} & 0.695 {\tiny \textcolor{red}{(-.042)}}  & 0.588 {\tiny \textcolor{red}{(-.024)}} & 0.605 {\tiny \textcolor{red}{(-.050)}} & 0.590 {\tiny \textcolor{green}{(+.002)}} & 0.592 {\tiny \textcolor{red}{(-.027)}} \\
\textbf{+ Joint Train} & 0.779 {\tiny \textcolor{green}{(+.084)}}  & 0.641 {\tiny \textcolor{green}{(+.053)}} & 0.703 {\tiny \textcolor{green}{(+.098)}} & 0.644 {\tiny \textcolor{green}{(+.054)}}  & 0.658 {\tiny \textcolor{green}{(+.066)}} \\
\rowcolor[gray]{0.95} \textbf{+ Token Init.} & 0.832 {\tiny \textcolor{green}{(+.053)}} & 0.690 {\tiny \textcolor{green}{(+.049)}} & 0.878 {\tiny \textcolor{green}{(+.175)}} & 0.652 {\tiny \textcolor{green}{(+.008)}} & 0.743 {\tiny \textcolor{green}{(+.085)}} \\
\textbf{+ Visual P.} & 0.878 {\tiny \textcolor{green}{(+.046)}} & 0.723 {\tiny \textcolor{green}{(+.033)}} & 0.890 {\tiny \textcolor{green}{(+.012)}} & 0.658 {\tiny \textcolor{green}{(+.006)}}  & 0.754 {\tiny \textcolor{green}{(+.011)}} \\
\rowcolor[gray]{0.95} \textbf{+ AL} & 0.899 {\tiny \textcolor{green}{(+.021)}} & 0.748 {\tiny \textcolor{green}{(+.025)}} & 0.914 {\tiny \textcolor{green}{(+.014)}} & 0.671 {\tiny \textcolor{green}{(+.013)}}  & 0.765 {\tiny \textcolor{green}{(+.011)}} \\
\bottomrule
\end{tabular}
\vspace{-6mm} 
\end{table}  

\noindent \textbf{Design of Module.}
As shown in Tab.~\ref{tab:module},  we sequentially evaluate our three core techniques: joint training, texual token initialization, visual prompting and auxiliary loss. Incorporating HNS significantly improves concept Rec, VG, choice-based tasks, and captioning, while VQA remains less affected—likely due to BLEU’s sensitivity to text. Overall, these techniques yield substantial gains, with visual prompting delivering the largest improvements in Rec and VG, and joint training better preserving language generation in VQA compared to direct parameter concatenation.

\noindent \textbf{Performance on Visually Similar Concepts.}
To evaluate the robustness of MC-LLaVA in challenging scenarios, we constructed a dedicated test set consisting of 5 highly visually similar scenarios, detailed in Sec.~\ref{sec:new_dataset}. As shown in the two leftmost images of Fig.~\ref{fig:ablation2}, RAP-MLLM and PeKit exhibit significant performance degradation. This failure stems from the inherent limitation of pure retrieval mechanisms, which struggle to distinguish fine-grained features among visually similar concepts. In contrast, our learning-based approach, through joint training and mask-based supervision, enables the model to capture subtle discriminative details. Consequently, MC-LLaVA achieves superior performance in disambiguating concepts, demonstrating the necessity of our training-based strategy for real-world applications.

\noindent \textbf{Generalizability across Backbones.}
To evaluate the architectural universality of MC-LLaVA, we adapt our framework to various state-of-the-art VLMs, including Qwen2.5-VL~\cite{bai2025qwen2} and LLaVA-OneVision~\cite{li2024llava}. As shown in the right part of Fig.~\ref{fig:ablation2}, our method consistently yields performance gains across both Rec and VQA tasks, regardless of the underlying model architecture. The stable improvements observed across diverse backbones demonstrate that MC-LLaVA's joint training and prompting strategies are not overfitted to a specific model but possess strong generalizability and robustness for VLM personalization.

We conduct additional ablation study on \textbf{Concept Token Length}, \textbf{Metrics} and \textbf{Initial Method} in the Appendix.

%% file: sec/6_conclusion.tex
\section{Conclusion}

We present MC-LLaVA, a novel multi-concept personalized vision-language model that significantly improves accuracy and efficiency via multi-concept instruction tuning equipped with personalized textual prompt and personalized visual prompt. 
Furthermore, we propose an auxiliary loss that effectively bridges personalized textual and visual prompts, pushing the model's performance upper bound. 
Our work not only advances frontiers of VLM personalization but also offers a high-quality multi-concept instruction dataset for future research. 
MC-LLaVA’s excellent performance across multiple tasks among  
various 
benchmarks highlights its ability to generate personalized responses
based on user-provided concepts. With growing demand for personalized services, MC-LLaVA and its 
dataset provide a strong foundation for developing more intelligent and user-specific assistants. 
This advancement further paves the way for new opportunities in real-world applications and transformed how we interact with assistants. 

%% file: sec/X_suppl.tex
\clearpage

\section{Catastrophic Forgetting}

Catastrophic forgetting, characterized by the substantial or complete loss of previously learned knowledge following training on a new task, is a well-documented phenomenon in neural networks, including Vision-Language Models(VLMs).
To quantify the impact of catastrophic forgetting in MC-LLaVA, we conducte a comparative analysis against the Original LLaVA-1.5-13B~\cite{liu2024visual} across established these multimodal benchmarks: MM-bench~\cite{li2023evaluating}, POPE~\cite{liu2024visual}, LLaVA-Wild~\cite{liu2025mmbench}.
The results are detailed in Tab.~\ref{tab:cas}.
Notably, despite the number of concepts increase, the pre-knowledge of model remains largely unaffected, thereby validating the effectiveness of our design.

\begin{table*}[h]
\centering
\begin{minipage}{0.49\textwidth}
\centering
\setlength{\tabcolsep}{2.5pt}
\begin{tabular}{l|ccc|c|c}
\toprule
\multirow{2}{*}{\textbf{Benchmark}} & \multicolumn{3}{c|}{\textbf{POPE}} & \textbf{MMBench} & \textbf{LLaVA-Wild} \\
& rand & pop & adv & en & \\
\midrule
\textbf{LLaVA} & 0.87 & 0.87 & 0.86 & 0.68 & 72.3 \\
\rowcolor[gray]{0.95}\textbf{MC-LLaVA}  & 0.86 & 0.86 & 0.85 & 0.67 & 72.2 \\
\bottomrule
\end{tabular}
\vspace{3mm}
\caption{\textbf{Catastrophic forgetting evaluation.} Results reveal that MC-LLaVA maintains same performance compared with Vanilla LLaVA, while enabling multi-concept personalized conversations.}
\label{tab:cas}
\end{minipage}
\hfill
\begin{minipage}{0.49\textwidth}
\centering
\begin{tabular}{l|c|c}
\toprule
\textbf{GPT-Score} & \textbf{Yo'LLaVA-M} & \textbf{MC-LLaVA} \\
\midrule
\textbf{Tokens}   & $10^1$ & $10^1$ \\
\midrule
\textbf{Single}   & 0.488 & 0.564 \\
\textbf{Multi}    & 0.389 & 0.602 \\
\textbf{Weighted} & 0.463 & 0.577 \\
\bottomrule
\end{tabular}
\vspace{3mm}
\caption{\textbf{Comparison of captioning capability between MC-LLaVA and Yo'LLaVA-M.} Scores are scaled from 0$\sim$10 to 0$\sim$1.}
\label{tab:caption}
\end{minipage}
\end{table*}

\vspace{-4mm}

\section{Discussion}

\vspace{-1mm}

\subsection{The Effect of VLMs' Prior Knowledge}
\begin{figure}[h]
    \centering
\includegraphics[width=0.6\linewidth]{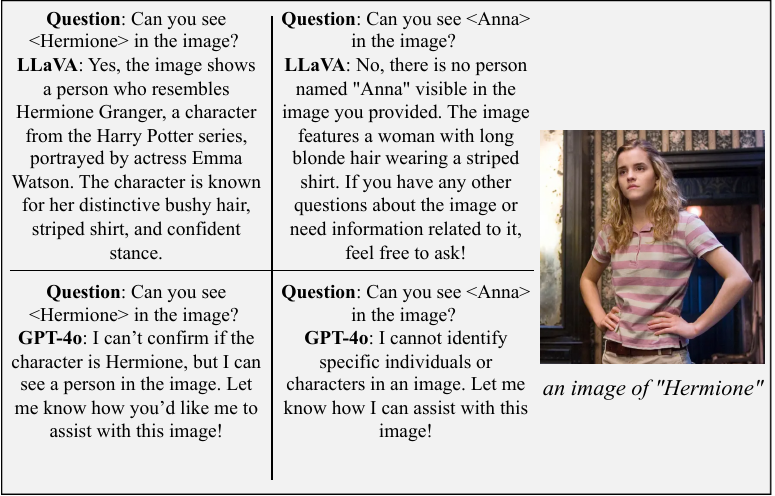}
    \caption{\textbf{Example conversation with GPT-4o and LLaVA on identifier \textless Hermione \textgreater and a random name \textless Anna\textgreater.}}
    \label{fig:her}
\end{figure}

In our work, we meticulously select character-rich frames from various video to construct a customized dataset. 
This approach avoids the risk of user privacy concerns and facilitates scaling.
However, there is a potential issue that the Visual Language Model itself may have prior knowledge of certain concepts.
Consequently, we conducted an in-depth exploration and discussion.
Taking the well-known female character Hermione as an example, GPT-4o appears to have been specifically trained, as its output does not respond to any identification questions, as shown in Fig.~\ref{fig:her}.

LLaVA might recognize \textless Hermione\textgreater; however, once the character’s identifier is personalized to a random name, such as \textless Anna\textgreater, LLaVA can no longer recognize this newly introduced user-defined concept. This naming approach aligns with how our dataset assigns names to concepts, which prevents the model from directly leveraging pre-trained knowledge for inference.
The vanilla LLaVA's poor captioning performance(close to 0) in the Tab.~\ref{exp:rec} of main text also reflect that even if a VLM has encountered a concept during pre-training, while we personalize it in downstream tasks, its pre-training may not offer significant assistance. 
Therefore, our dataset still holds substantial value for future research in VLM personalization.

\subsection{The Addtional Assessment of Captioning}
In our main experiments evaluating captioning capabilities, we strictly follow the experimental setup outlined in MyVLM~\cite{alaluf2025myvlm}. 
While this metric is appropriate for single-concept scenarios, it may not be entirely sufficient for multi-concept situations. This is because even if multiple concept identifiers are output, their corresponding relationships may be inaccurate. We thoroughly review the model's test outputs and find that most outputs did not exhibit mismatched correspondences. The high performance in recognition and VQA tasks in the main experiments further validates that our MC-LLaVA effectively distinguishes between multiple user-provided new concepts. 

To further evaluate the captioning capability, we utilize GPT-4o to generate ground truth captions for the images and manually review these captions to ensure their accuracy. We employ GPT-4o to score the captions generated by MC-LLaVA and Yo'LLaVA-M against the ground truth across three dimensions: Accuracy, Helpfulness, and Relevance. The scores from these dimensions are then weighted and summed to obtain a final score, with a maximum of ten points and a minimum of zero points. The detailed prompt is shown in Fig.~\ref{fig:gpt4o-cap-template} and the results of the GPT-based scoring are summarized in the Tab.~\ref{tab:caption}.



\section{Additional Ablations}
\subsection{The Number of Trainable Concept Tokens}
\begin{figure}[t]
    \centering
\includegraphics[width=\linewidth]{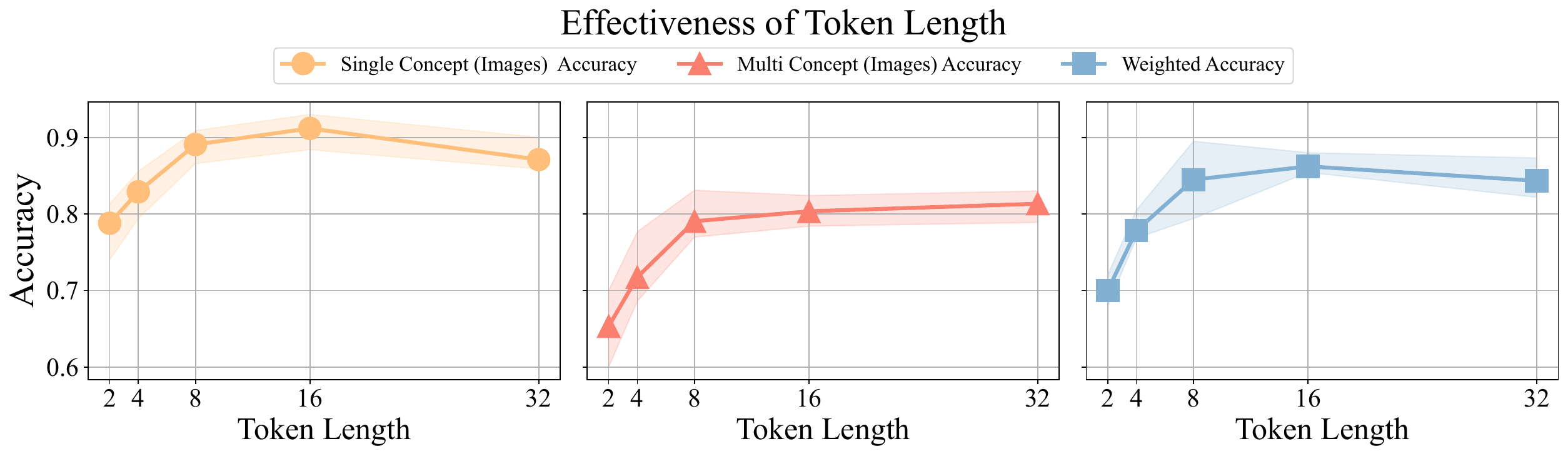}
    \caption{\textbf{Recognition performance comparison of MC-LLaVA under different numbers of tokens per concept.}}
    \vspace{-5mm}
    \label{fig:ablation_token}
\end{figure}

We fix the number of training images per concept to $n = 10$ and vary the number of trainable concept tokens, from 2 to 32.
As illustrated in Fig.~\ref{fig:ablation_token}, increasing the length of trainable tokens enhances the model's recognition ability for both single and multiple concepts, especially when the token length exceeds 8.
Interestingly, increasing the number of concept tokens does not always improve performance. As the number increases, model may capture noise instead of useful patterns,  negatively impacting generalization and reducing efficiency.

\subsection{Time and Space Overhead Analysis}
We measured the average inference time per conversation and peak memory usage across all evaluation experiments. As summarized in Tab.~\ref{tab:overhead}, our approach introduces only marginal additional latency and memory usage compared to Yo'LLaVA in 2-concept scenarios. When scaling to more concepts, the overhead increases slightly but remains negligible.

\subsection{The Comparison of Texual Token Initialization Method}
We conducted experiments using the Random and PCA Initialization methods as shown in Tab.~\ref{tab:comparison_init}. The results indicate that k-means consistently outperforms these methods, likely due to its ability to effectively capture the data's structure, leading to better convergence. While simple, k-means is effective and aligns with Occam's razor.

\begin{table*}[ht]
\centering
\begin{minipage}{0.48\textwidth}
\centering
\setlength\tabcolsep{1pt}
\renewcommand\arraystretch{1.3}
\begin{tabular}{l|cccc}
\toprule
 & \textbf{Yo'LLaVA} & \textbf{Ours w/o VP} & \textbf{Ours} &\textbf{ Ours (3C)} \\
\midrule
\textbf{Time (s)} & 0.7831 & 0.7833 & 0.8147 & 0.8399 \\
\textbf{Memory (MB)} & 31160 & 31160 & 31160 & 31161 \\
\bottomrule
\end{tabular}
\vspace{3mm}
\caption{\textbf{Comparison of inference time and memory usage on A800 GPUs.} VP = visual prompt; 3C = 3-concept scenario.}
\label{tab:overhead}
\end{minipage}
\hfill
\begin{minipage}{0.48\textwidth}
\centering
\setlength{\tabcolsep}{1pt}
\renewcommand{\arraystretch}{1.3}
\begin{tabular}{l|c|cccc}
\toprule
 & \textbf{Type}& \textbf{None} & \textbf{Pooling} & \textbf{PCA} & \textbf{K-means} \\ \midrule
\textbf{Rec.} & Weighted &0.842&0.837 &0.852 &  0.862 \\ 
\textbf{VQA} & Weighted &0.649 &0.617 &0.647 &0.652 \\ 
\bottomrule
\end{tabular}
\vspace{3mm}
\caption{\textbf{Comparison of different token initialization methods.} \newline}
\label{tab:comparison_init}
\end{minipage}

\end{table*}

\subsection{More Compared Baselines}  
To further strengthen our evaluation, we additionally assess our benchmark using supplementary methods and report the results in Tab.~\ref{tab:appendix_baselines}. MC-LLaVA consistently outperforms strong baselines (e.g., Qwen2.5-VL-7B and GPT-4V with handcrafted prompts) across personalized recognition (Rec.), visual grounding (VG), and VQA.

\subsection{Additional Metrics on VQA Task}  
As shown in Tab.~\ref{tab:appendix_metrics}, MC-LLaVA consistently surpasses Yo'LLaVA-S and Yo'LLaVA-M under complementary evaluation metrics on the VQA task in our benchmark, including METEOR, ROUGE-L, and BERTScore. This further demonstrates the robustness and effectiveness of MC-LLaVA in handling personalized VQA tasks.

\begin{table*}[ht]
\centering
\renewcommand\arraystretch{1.25}

\begin{minipage}{0.49\textwidth}
\centering
\begin{tabular}{l|ccc}
\toprule
\textbf{Method} & \textbf{Rec.} & \textbf{VG} & \textbf{VQA} \\
\midrule
\textbf{Qwen2.5-VL-7B + Prompt} & 0.627 & 0.606 & 0.518 \\
\textbf{GPT-4V + Prompt}        & 0.776 & 0.681 & 0.698 \\
\rowcolor[gray]{0.95} \textbf{MC-LLaVA} & \textbf{0.878} & \textbf{0.723} & \textbf{0.679} \\
\bottomrule
\end{tabular}
\vspace{3mm}
\caption{\textbf{Comparison with additional baselines.}}
\label{tab:appendix_baselines}
\end{minipage}
\hfill
\begin{minipage}{0.49\textwidth}
\centering

\begin{tabular}{l|ccc}
\toprule
\textbf{Method} & \textbf{METEOR} & \textbf{ROUGE-L} & \textbf{BERTScore} \\
\midrule
\textbf{Yo'LLaVA-S} & 0.471 & 0.612 & 0.875 \\
\textbf{Yo'LLaVA-M} & 0.459 & 0.585 & 0.866 \\
\rowcolor[gray]{0.95} \textbf{MC-LLaVA} & \textbf{0.482} & \textbf{0.633} & \textbf{0.889} \\
\bottomrule
\end{tabular}
\vspace{3mm}
\caption{\textbf{Results on additional metrics on the VQA task.}}
\label{tab:appendix_metrics}
\end{minipage}

\end{table*}


\section{Related Work}
\noindent \textbf{Personalized VLMs.}
The emergence of LLMs has revolutionized the way humans interact with machines.
Thanks to their superior capabilities, LLMs serve as fundamental building blocks for our daily lives~\cite{chen2024large,croft2001relevance,salemi2023lamp}.
To enhance individual experiences and preferences, personalized LLMs consider user personas to better meet customized needs. 
However, in the context of VLMs, personalized models require not only textual information but also additional visual information to aid in understanding concepts. 
Although recent works~\cite{alaluf2025myvlm,nguyen2024yo, pi2024personalized} have begun to explore VLM personalization, they primarily focus on single-concept scenarios. 
While recent RAG-based~\cite{hao2024remember} and training-free~\cite{seifi2025personalization} approaches have made progress in personalization, they often struggle with multi-concept scenarios, particularly when faced with high visual similarity among reference samples. In contrast, our work provides a systematic investigation and comparative analysis of multi-concept personalization, demonstrating superior performance in disambiguating visually similar concepts.

\par \textbf{Vision Language Models.}
Recently, LLMs have achieved significant advancements~\cite{achiam2023gpt,touvron2023llama,bai2023qwen, yang2023baichuan}. Following this, the emergence of VLMs~\cite{liu2024visual,lin2023video, 
bai2023qwenvl,lin2023sphinx} has significantly expanded the capabilities of LLMs, enabling them to be applied in tasks such as data processing~\cite{luo2024llm}, VQA~\cite{guo2023images}, and captioning~\cite{li2023blip}. Although they exhibit strong general abilities in numerous tasks, it is challenging for them to fulfill the requirements for highly personalized responses~\cite{alaluf2025myvlm}, especially when multiple concepts are involved. In this work, we propose a multi-concept personalization method, allowing VLMs to reason accurately with multiple concepts.

\noindent \textbf{Parameter-Efficient Fine-Tuning.}
LLMs and VLMs excel in a wide range of downstream tasks. However, updating and storing all model parameters for each task has become increasingly expensive. Compared to re-train the whole model, Parameter-Efficient Fine-Tuning (PEFT) methods~\cite{hu2021lora, zhang2024gradient, houlsby2019parameter, zhang2023split, zhang2023mosa} achieves training and storage efficiency by updating only a small subset of parameters. Among PEFT, prompt tuning~\cite{lester2021power,liu2021p, jia2022visual} is one of the most widely used methods. 
Prompt tuning primarily involves manually designed hard prompts~\cite{wen2024hard} and learnable soft prompts~\cite{li2021prefix,zhou2022conditional}. 
While soft prompt tuning has achieved notable successes across various tasks, its effectiveness depends on the appropriate initialization of parameters~\cite{meng2024pissa}, which leads to current personalization approaches heavily rely on the availability of high-quality negative samples~\cite{nguyen2024yo}.
In this paper, we propose a simple yet effective approach for initializing concept tokens, which reduces reliance on negative samples. 
Additionally, we find that this method effectively accelerates the convergence speed.

\section{Multi-Concept Instruction Dataset}
Our dataset includes nearly 2,000 images. There are 10 training images for each concept, 5 single-concept images and 5 extra multi-concept scenario images, which belong to two, three, and four multi-concept scenarios. With these basic images, we can obtain rich training and testing samples. Below, we will specifically show the detailed training samples of all the above-mentioned types.

\subsection{Training Data Explanation}
To train MC-LLaVA, we need to construct training samples. We leverage a unified training data form ($I$,$X_q$,$X_a$), where $I$ is the input image, $X_q$ is the question, and $X_a$ is the answer. We collect training samples from the following tasks:
\begin{itemize}[noitemsep, topsep=0pt, leftmargin=*]
\item \textbf{Positive Recognition}: To better integrate concepts into VLMs, we adopt a positive recognition task following Yo'LLaVA, assigning multiple positive recognition conversations to each concept image. 
\item \textbf{Random Recognition}: To avoid repetitive ``Yes'' responses from the model, we randomly select 100 images from CC12M~\cite{changpinyo2021conceptual} as inputs for the negative recognition task. These images are paired with conversations generated from a negative template, eliminating the need for visually similar images, which are convenient to collect.
\item \textbf{Joint Recognition}: Joint training not only helps in acquiring effective negative samples for a specific concept but also improves the model's ability to distinguish between concepts through inter-concept negative sampling. Specifically, images from $\langle \text{sks}_1 \rangle$ serve as input while negative  conversations are generated using $\langle \text{sks}_2 \rangle$ from the negative template. This approach allows for generating at least $m\times(m-1)\times n$ negative samples, given $m$ concepts with $n$ images each.
\item \textbf{Conversation}: Recognition tasks alone do not adequately prepare the model for conversational proficiency; thus, incorporating standard QA samples is crucial. For each concept with n images, we create 10 consistent general questions focusing on visual attributes, with answers provided by GPT-4o. Notably, while Yo'LLaVA utilizes text-only dialogues for training, we notice that the same concept can exhibit different visual features, such as hairstyles, across various images. Ignoring image information can lead to inconsistencies in quality assurance responses. Furthermore, our experiments indicate that VQA performs better without affecting the model's effectiveness in text-only conversations. 
\end{itemize}

\section{Experiment}
\subsection{Baselines}

We supplement the baselines not described in the main text:

\begin{itemize}[noitemsep, topsep=0pt, leftmargin=*]
    \item \textbf{LLaVA}: Vanilla LLaVA~\cite{liu2024visual} without any personalized information.
    \item \textbf{LLaVA+Prompt}: We first prompt LLaVA to generate captions for all training images of a concept, then use these personalized captions in two ways: (I) concatenate all captions to form a comprehensive description of the concept; and (II) prompt LLaVA to summarize the captions into a concise, personalized description. During inference, we add $m$ relevant captions to the input to supply concept-specific information, where $m$ is the number of concepts evaluated.
    \item \textbf{GPT4o+Prompt}: Similar to LLaVA+Prompt, but using GPT-4o as the base model, which serves as an upper bound. Notably, the GPT4o employed for testing differs from that used for data generation to avoid potential knowledge leakage.
    \item \textbf{MyVLM}~\cite{alaluf2025myvlm}: We employ the MyVLM-LLaVA model. For multi-concept scenarios, additional concept heads are trained for each concept. Due to MyVLM's limitations, only one concept head is utilized during inference, preventing the model from addressing questions that involve multiple concepts simultaneously.
\end{itemize}

\subsection{Testing Task Component}
\noindent \textbf{Recognition} We consider a scenario consisting of $n$ concepts. 
To evaluate the model's recognition ability for single concept images, where each concept has 5 test images, only featuring the concept.
Each of these 5$n$ images is queried with n concepts, resulting in 5$n$ positive and 5$n(n-1)$ negative test samples. 
Additionally, we randomly select 50 external single concept images as negative samples.
For multi-concept image recognition, each scenario includes 5 test images, where each contains up to $n$ concepts. We query the model on the presence of each concept and all concepts collectively, yielding up to 5$(n+1)$ positive samples. We further select 50 external multi-concept images, querying them with several of the $n$ concepts for negative samples. In total, the method is assessed on up to 5$n^2$ + 5$(n+1)$ + 100 recognition tasks, comprising 5$n$ + 5$(n+1)$ positive and 100 + 5$n(n-1)$ negative samples. In total, we utilized 3,155 and 2,665 test samples for the single-concept and multi-concept scenarios, respectively. 

\noindent \textbf{Question Answering} All visual tasks share the same data composition. For a scenario with $n$ concepts, each of the 5 single concept test images contributes 5 QA pairs, resulting in 5$n$ QA pairs for single concept images. Each of the 5 multi-concept images, assuming they contain all n concepts, generates $2^n - 1$ QA pairs (corresponding to the non-empty subsets of an n-element set). To sum up, for each task, a scenario with n concepts can have up to $5(n+2^n-1)$ QA pairs for testing.
For text-only QA, an n-concept scenario contains $5n$ single-concept QA pairs and 5 multi-concept QA pairs, resulting in $5n+5$ text-only QA pairs for testing.

\noindent \textbf{Captioning}
In an n-concept scenario, all test images—comprising 5$n$ single-concept images and 5 multi-concept images—are utilized for concept experiments. We prompt the model to generate captions for each image and quantitatively assess the model's captioning capability based on the presence of identifiers within the images. This provides a metric for evaluating the model's ability to accurately caption and recognize concepts in both single and multi-concept settings.

\subsection{Implementation Details}


We use LLaVA-1.5-13B~\cite{liu2024visual} as the VLM backbone for all experiments. During training, the $n$ concepts within a single scenario are jointly trained in one pass. For each concept, all 10 images from the training set are used, with a batch size of 1. Training is conducted for 15 epochs, and at the end of the final epoch, we save the token embeddings corresponding to each concept and the parameters of the LM head.
Empirically, we set $\lambda = 0.2$, $\tau = 0.32$, $\gamma = \frac{100}{256 \times 256}$ and $\beta = 0.5$.
The training process consists of two phases. First, we initialize the parameters using $k$-means clustering~\cite{hartigan1979algorithm} with Euclidean distance. Then, during the VQA-based training phase, we optimize the model using AdamW~\cite{kingma2014adam} with a learning rate of 0.001, applying the standard masked language modeling loss. Each epoch consists of an average of $250 + 100(n-1)$ steps per concept.
For testing, due to the stochastic nature of language model generation and the randomness in test set composition (especially for recognition), each test task is performed three times using different seeds for all scenarios. The average result across the runs is reported.
All experiments are conducted on 80GB A100 GPUs.

We find that concept token initialization is crucial as misalignment with the tokenizer's embedding distribution can destabilize training. We normalize the vector norms of concept tokens \( K^1, \ldots, K^k \) (denoted \( K_* \)) from k-means. To align with tokenizer embeddings \( K_o \), adjusted tokens are:

\begin{equation}
\hat{K_*} = \frac{K_*}{||K_*||} \cdot K_o
\end{equation}


\subsection{Captioning Questions Template}
\label{sec:12.4}
We only use the image captioning question in the test phase. For single-concept captioning and multi-concept captioning, we use the same template, that is, ``Can you see \textless $\text{concept}_1$\textgreater \ldots \textless $\text{concept}_m$\textgreater \:in the image? Don't answer the question, but remember it, and only respond with a detailed caption for the image. Your caption:".

\subsection{Additional Qualitative Results}
\begin{figure*}[ht]
    \centering
\includegraphics[width=1\linewidth]{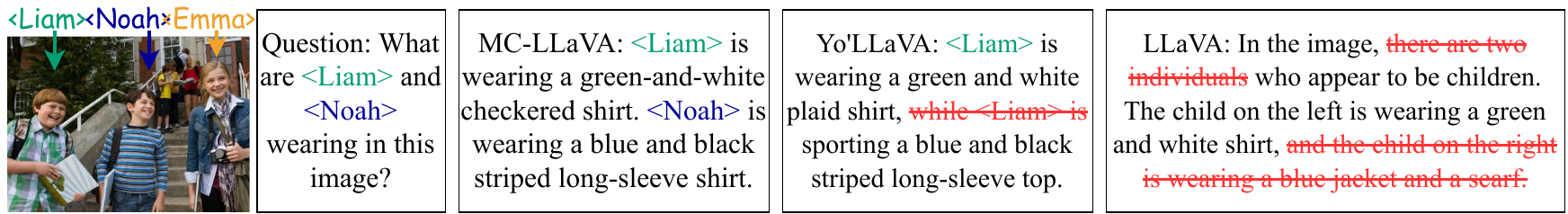}
        \vspace{-5mm}
    \caption{\textbf{The performance of the scenario displayed in the banner across different models.}}
    \label{fig:additional_com}
\end{figure*}

\begin{figure*}[ht]
    \centering
\includegraphics[width=1\linewidth]{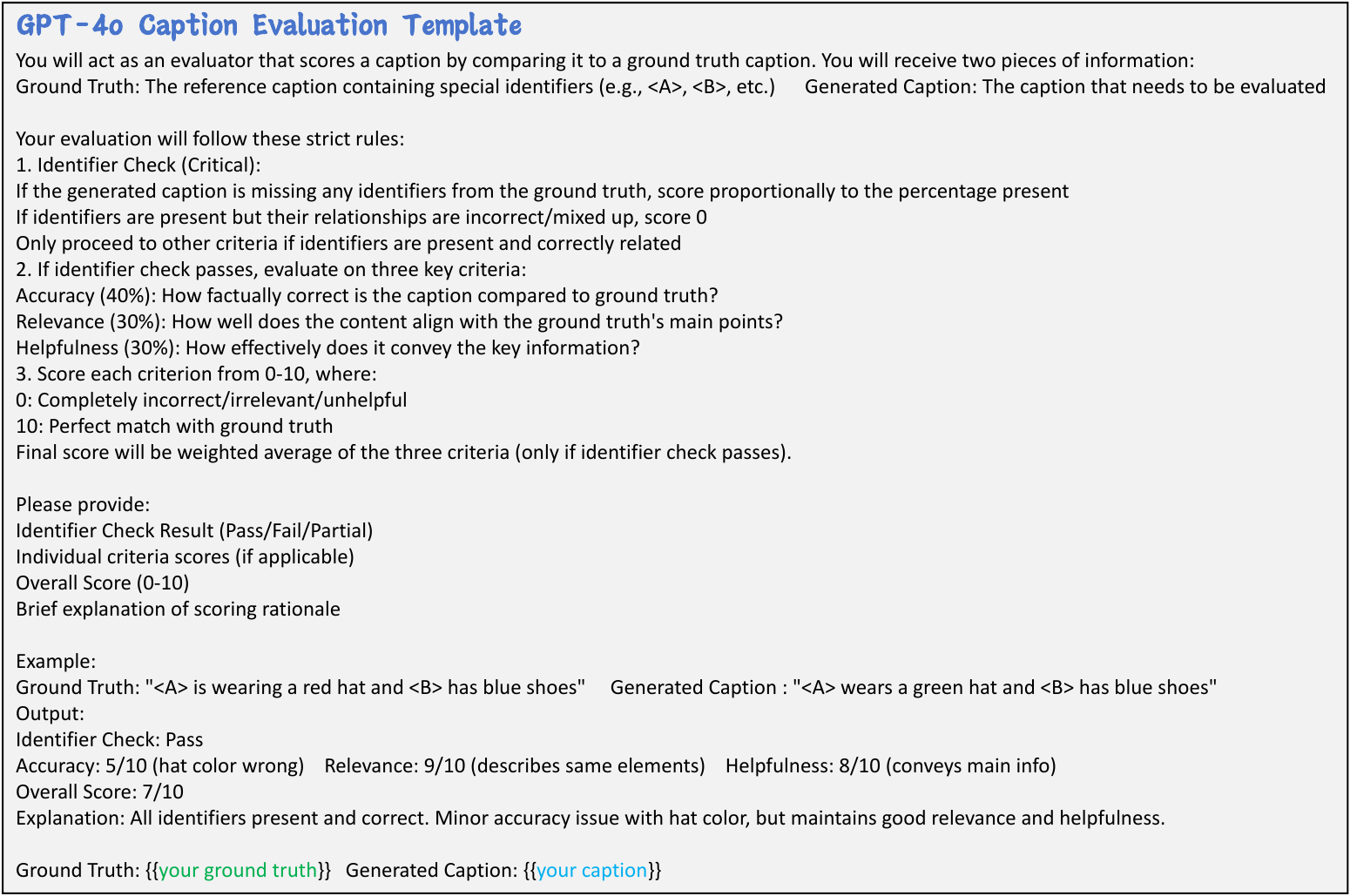}
        \vspace{-5mm}
    \caption{\textbf{The caption evaluation template prompted to GPT-4o}}
    \label{fig:gpt4o-cap-template}
\end{figure*}

\begin{figure*}[ht]
    \centering
    \includegraphics[width=0.8\textwidth]{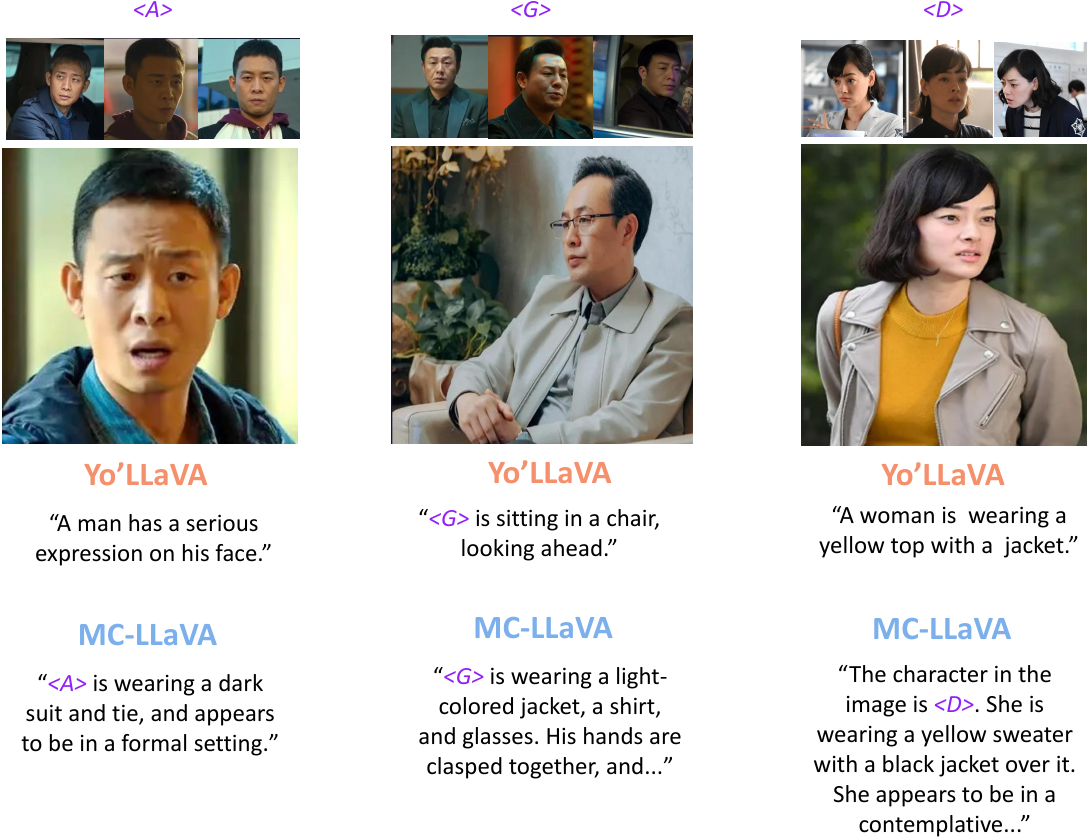}
    \caption{\textbf{Comparison of personalized captioning between Yo'LLaVA and \shortname.}}
    \label{fig:cap1}
    \vspace{-5mm}
\end{figure*}

\begin{figure*}[ht]
    \centering
    \includegraphics[width=0.8\textwidth]{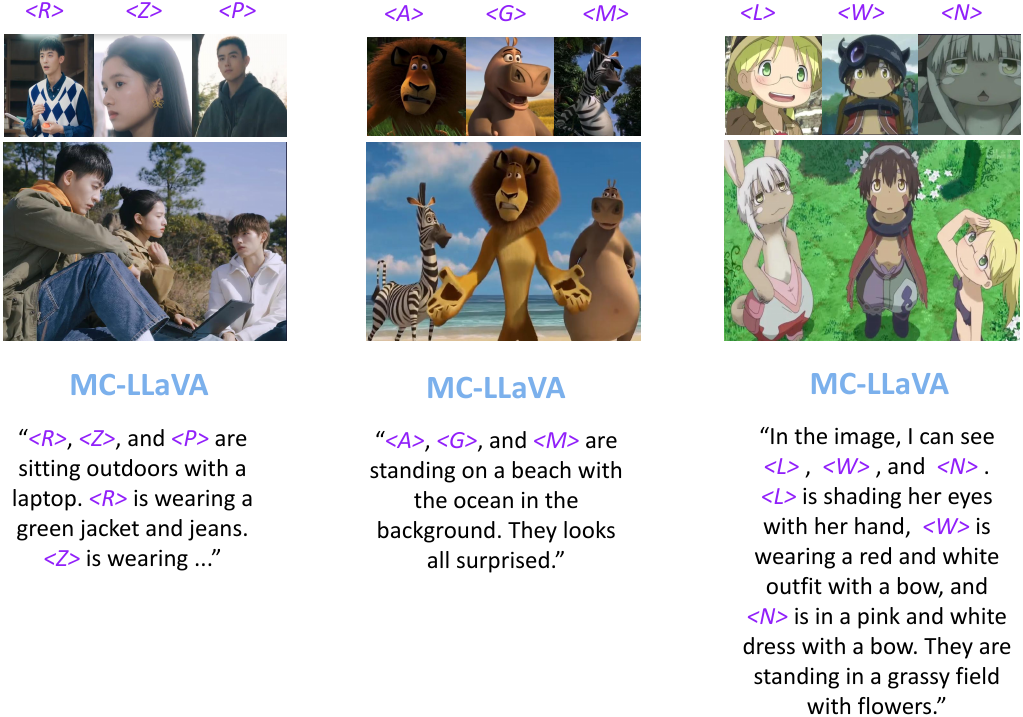}
    \caption{\textbf{Personalized caption of multi-concept with \shortname.}}
    \label{fig:cap2}
    \vspace{-5mm}
\end{figure*}

We provide qualitative results for visual question answering, image captioning, and multiple-choice questions, as follows:
\begin{enumerate}
    \item In Fig.~\ref{fig:additional_com}, we demonstrate the performance of different models in 3-concept scenarios from the banner Fig.~\ref{fig:fig1}.
    \item In Tab.~\ref{tab:herry_hermione_ron}, we provide examples of VQA in the three-concept scenarios.
    \item In Fig.~\ref{fig:cap1}, we compare the personalized captions of Yo'LLaVA~\cite{nguyen2024yo} and MC-LLaVA for single-concept scenarios. Both use the captioning questions template provided in Sec.~\ref{sec:12.4}.
    \item In Tab.~\ref{fig:cap2}, we show the personalized captions of MC-LLaVA in multi-concept scenarios, using the captioning questions template provided in Sec.~\ref{sec:12.4}.
\end{enumerate}

\section{Limitation and Future Work}
\begin{figure*}[t]
    \centering
    \includegraphics[width=1\linewidth]{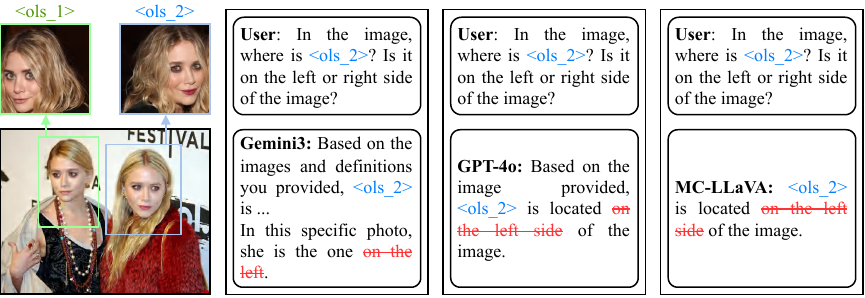}
    \caption{\textbf{Showcase of the performance of existing models in a scenario with concepts of extremely similar visuals.}}
    \label{fig:twins}
\end{figure*}

MC-LLaVA enhances the capacity for personalized interactions in vision-language models, particularly excelling in multi-concept scenarios. 
However, it is essential to acknowledge several limitations, which can serve as future directions.
Firstly, while MC-LLaVA leverages visual information to facilitate the accurate and efficient integration of new concepts into VLMs, the current process still necessitates training, which poses certain challenges for real-world deployment.
A promising avenue for future research is to explore the possibility of integrating new concepts to the model without training.
Secondly, while our multi-concept dataset pioneers task-level evaluation in VLM personalization, the field lacks comprehensive benchmarks that encompass a larger scale and capability dimensions.
This limitation restricts our assessment of capability-level VLM personalization. Future work could define the capabilities that models should be evaluated on and propose more comprehensive benchmarks that assess various aspects of personalization capabilities in VLMs.
Finally, as illustrated in Fig.~\ref{fig:twins}, scenarios composed of visually near-identical concepts (e.g., identical twins) remain challenging: even humans, MC-LLaVA, and strong proprietary models such as Gemini~3 and GPT-4o often fail to reliably distinguish them, leading to concept confusion. Addressing such inherently ambiguous cases remains an important open challenge for future work.

\begin{table*}[ht]
\renewcommand{\arraystretch}{2}
\setlength{\tabcolsep}{0.5mm}
  \begin{minipage}{1\textwidth}
\centering  
\scalebox{0.9}{
\begin{tabular}{l p{5.5cm} p{7.5cm} }
\toprule
 \multicolumn{3}{l}{\bf MC-LLaVA}  \\
\midrule
\multicolumn{3}{c}{
\texttt{<H>}:  \raisebox{-.4\height}{\includegraphics[height=1.5cm]{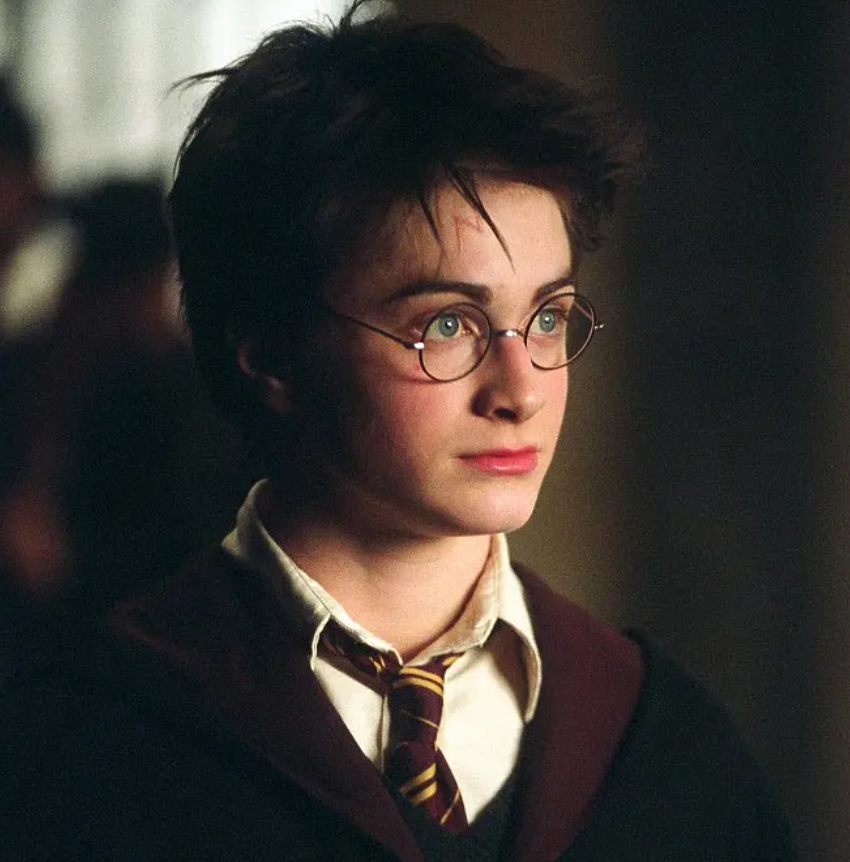}}
\texttt{<M>}:  \raisebox{-.4\height}{\includegraphics[height=1.5cm]{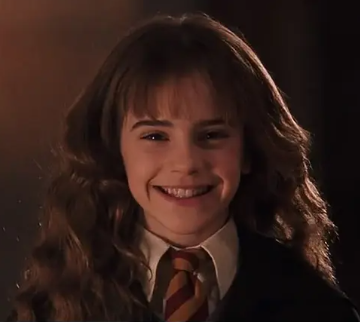}}
\texttt{<R>}:  \raisebox{-.4\height}{\includegraphics[height=1.5cm]{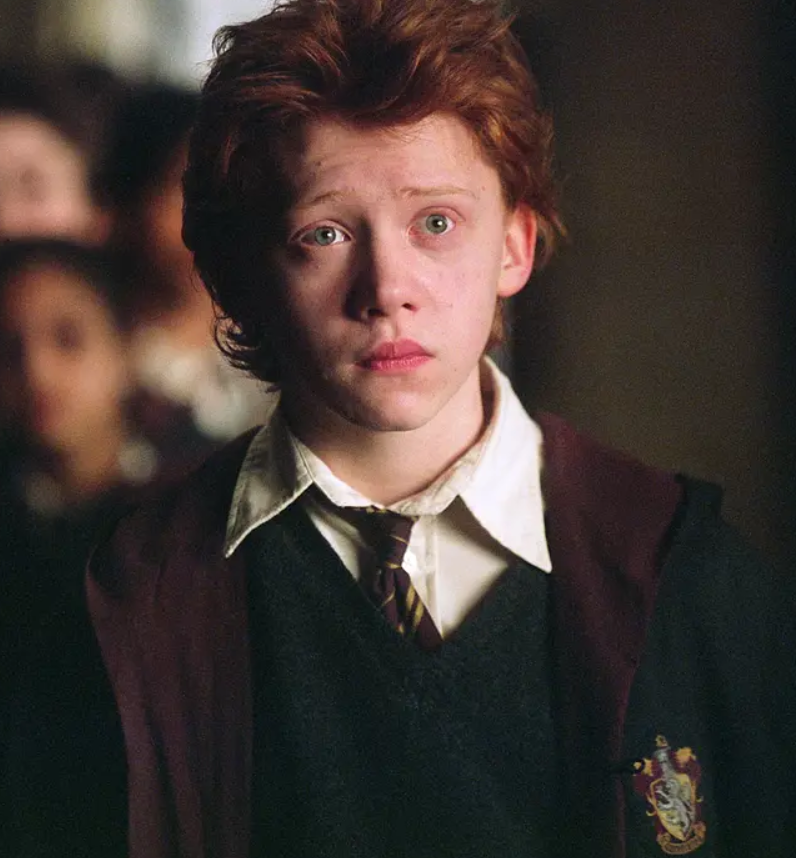}}
}
\\
\midrule
\multicolumn{3}{l}{\colorbox{mypink}{\textit{\textbf{$\triangleright$ Visual Conversation}}} --- (Note: Single-concept)} \\
&  \multicolumn{1}{c}{\includegraphics[height=3cm]{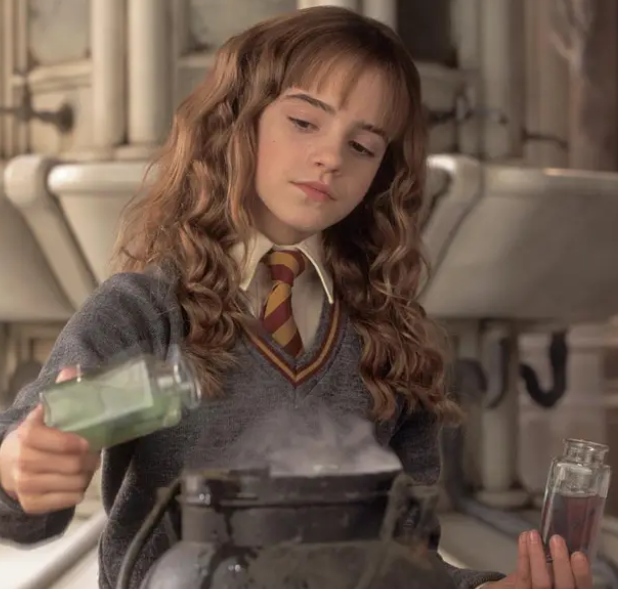}}
&  \multicolumn{1}{c}{\includegraphics[height=3cm]{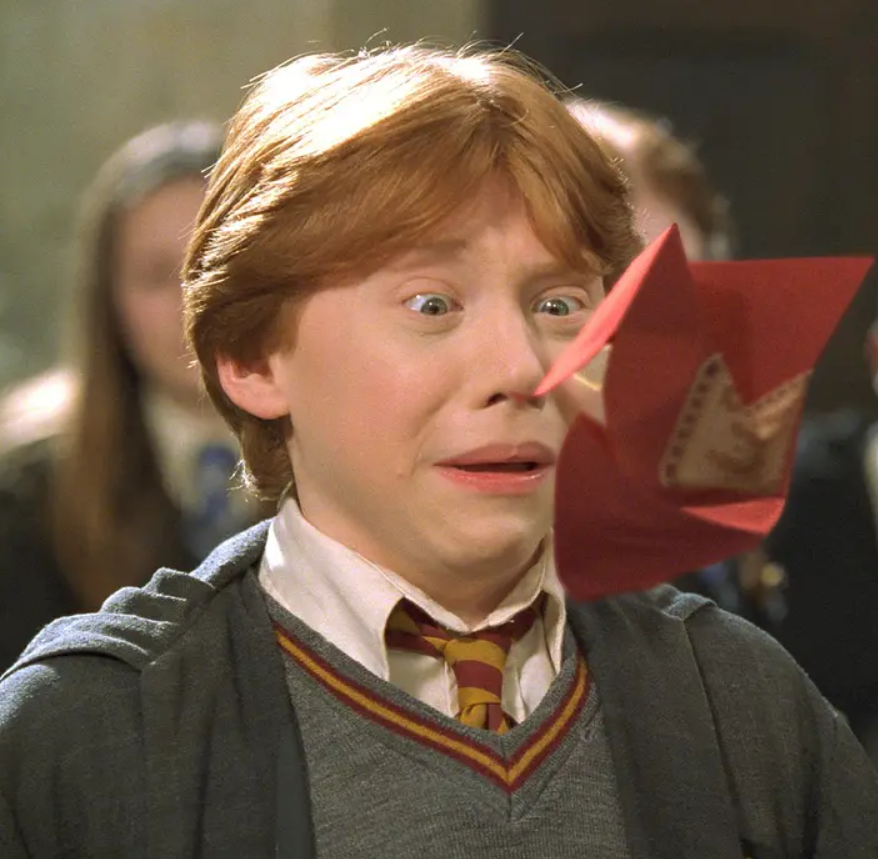}} \\
\midrule
Question:
& What is \texttt{<M>} doing?
& Is \texttt{<R>} in this photo? If yes, describe his location and the scenario.
\\
\midrule
\shortname:
& \texttt{<M>} is holding a green glass flask in her left hand and a clear glass flask in her right hand, suggesting she might be conducting an experiment or a study. 
& Yes, \texttt{<R>} is in this photo. \texttt{<R>} is standing in a classroom, his face contorted in an expression of shock. His eyes are wide and his mouth is slightly open, suggesting he has just been startled or caught off guard.
\\
\midrule
\multicolumn{3}{l}{\colorbox{mypink}{\textit{\textbf{	$\triangleright$ Visual Conversation}}} --- (Note: Multi-concept)} \\
&  \multicolumn{1}{c}{\includegraphics[height=3cm]{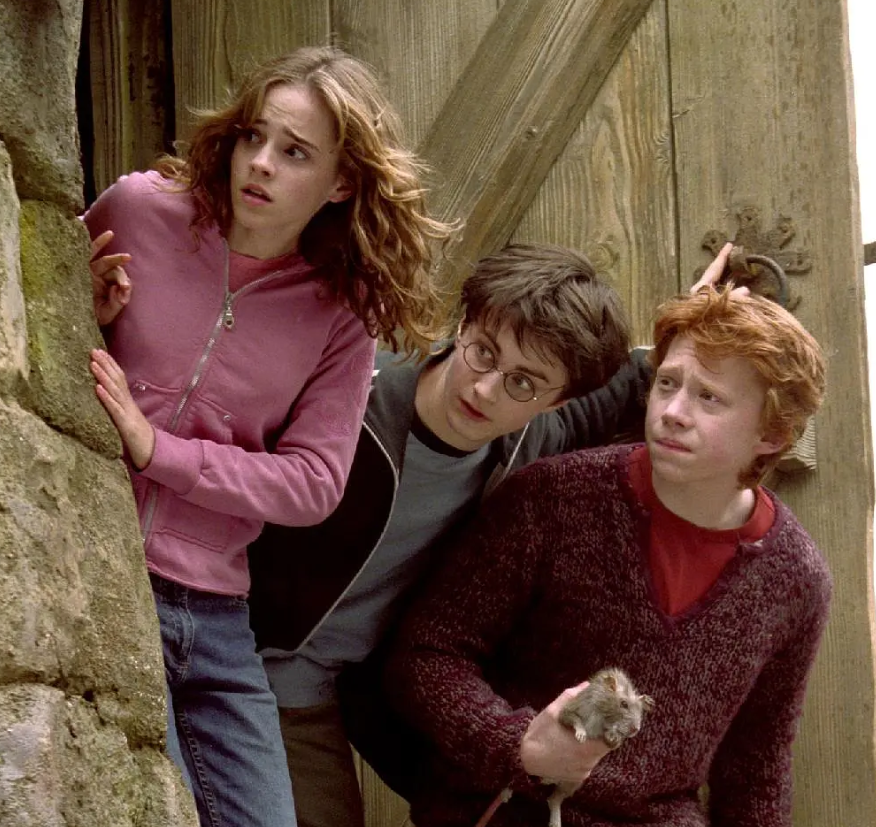}}
&  \multicolumn{1}{c}{\includegraphics[height=3cm]{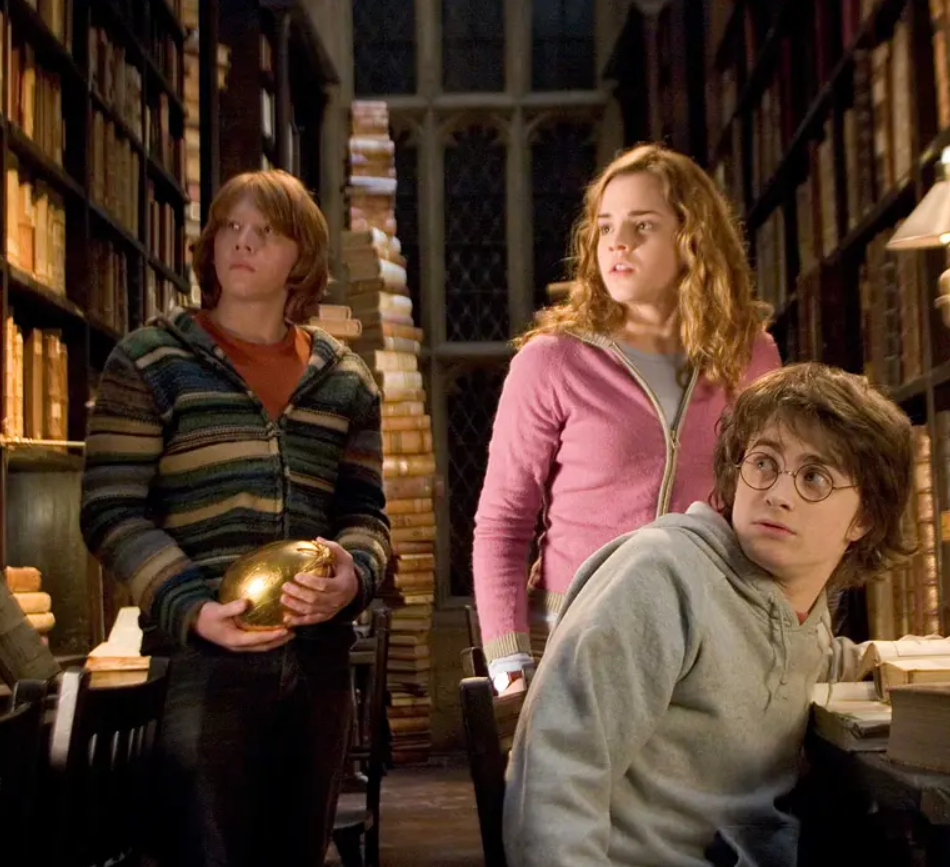}} \\
\midrule
Question:
& What are \texttt{<H>}, \texttt{<M>}, and \texttt{<R>} doing in this photo?
& Can you see \texttt{<H>} and \texttt{<M>} in this photo? Please describe the current scenario.
\\
\midrule
MC-LLaVA:
& \texttt{<H>} and \texttt{<M>} are peeking around a corner, with \texttt{<M>} wearing a pink hoodie. \texttt{<R>} is on the far right, holding a small animal in his hand.
& Yes, I can confirm that \texttt{<H>} and \texttt{<M>} are indeed in the photo. \texttt{<H>} is sitting at a table with books, while \texttt{<M>} is standing nearby in a library setting. \texttt{<R>} seems to be in the picture too, holding a golden egg in his hand.
\\
\bottomrule
\end{tabular}
}
\caption{\textbf{Examples of \shortname's abilities with a learned concept  \texttt{<H>}, \texttt{<M>} and \texttt{<R>}.}}
\label{tab:herry_hermione_ron}  
  \end{minipage}
\end{table*}

\section{Notes on the Visually Similar Dataset}

\label{sec:new_dataset}
\begin{figure*}[t]
    \centering
    \includegraphics[width=1\linewidth]{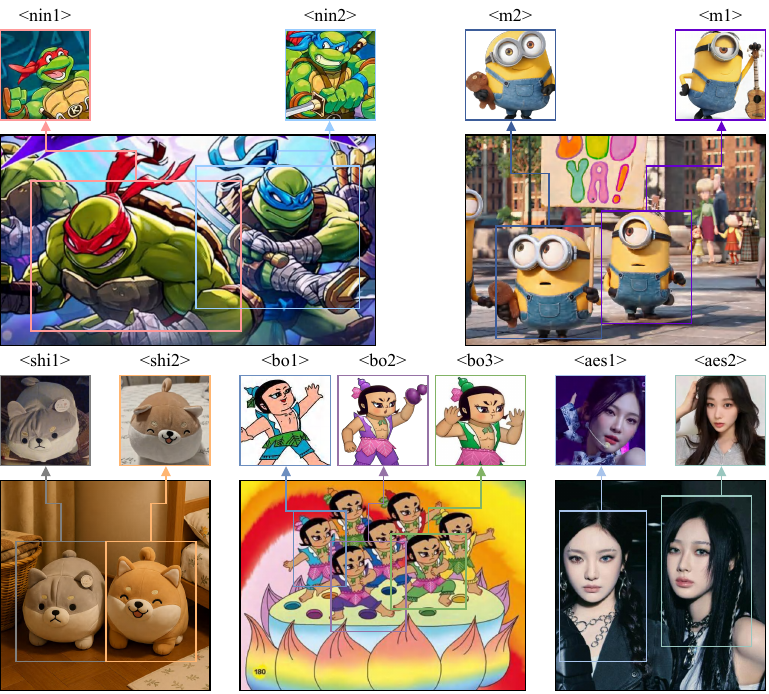}
    \caption{\textbf{Five newly collected multi-concept scenarios.}}
    \label{fig:iclr_rebuttal}
\end{figure*}


As shown in Fig.~\ref{fig:iclr_rebuttal}, we curate five multi-concept scenarios, each containing two or three concepts with similar visual attributes. These scenarios form a small toy dataset used to evaluate how well different models can distinguish between and remember visually similar concepts.
